\documentclass[11pt]{cuhksz}

\usepackage{cuhksz}
\usepackage{tabularx}
\usepackage{enumitem}
\usepackage{makecell}
\usepackage{wrapfig}

\newtheorem{desideratum}{Desideratum}

\newcommand{\benchmark}{\textbf{GameCraft-Bench}\xspace}
\newcommand{\modellogo}[1]{%
  \raisebox{-0.18em}{\makebox[1.35em][c]{%
    \includegraphics[height=1.05em,width=1.05em,keepaspectratio]{#1}}}%
}
\leftlogo{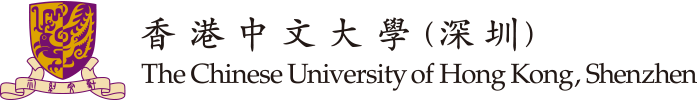}
\leftlogo{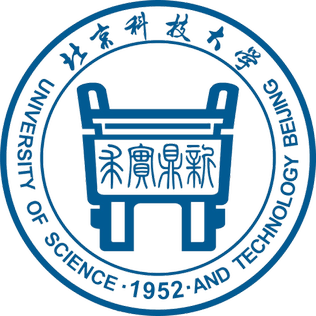}
\leftlogo{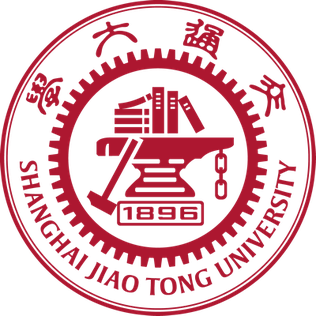}
\leftlogo{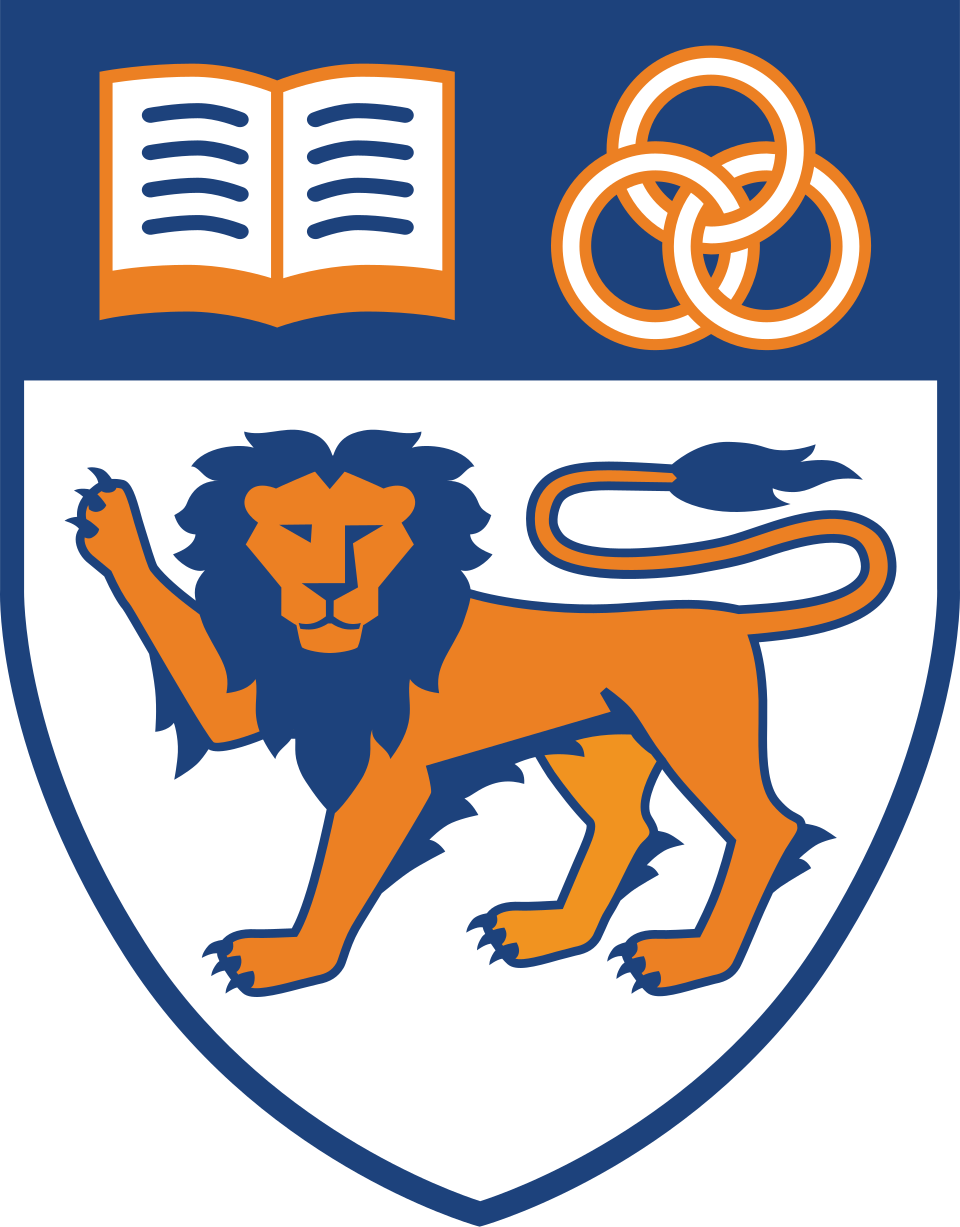}

\rightlogo{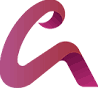}
\rightlogo{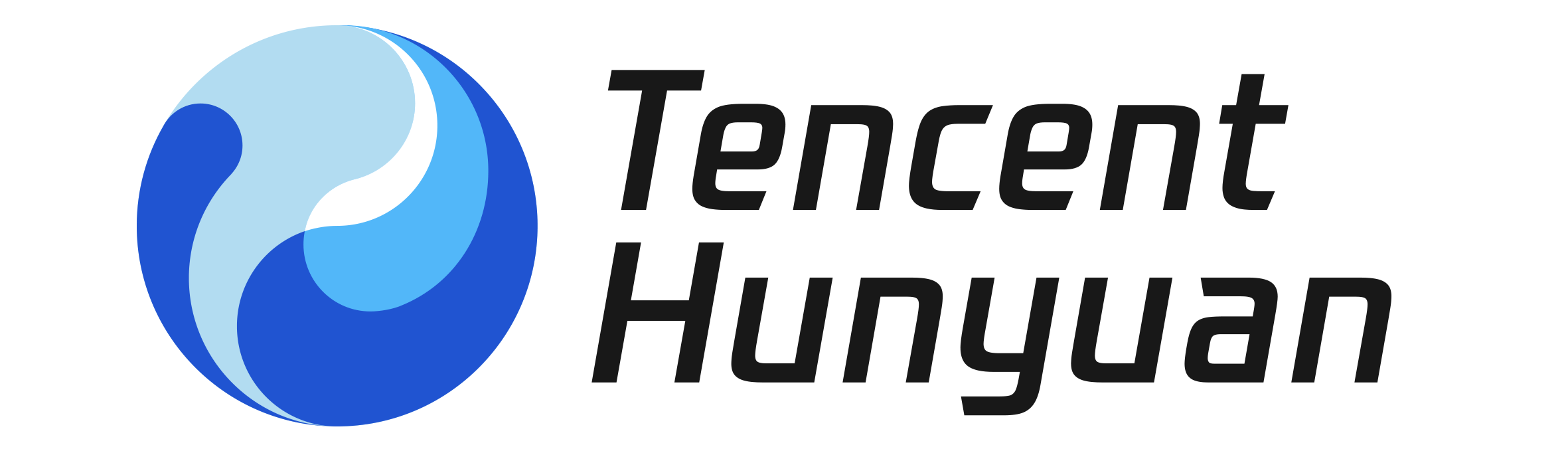}

\newtcolorbox{prompt}[1]{
    enhanced,
    left=4mm,
    right=4mm,
    top=2mm,
    bottom=2mm,
    boxsep=0mm,
    rounded corners,
    title=#1,
    fontupper=\footnotesize\linespread{0.9}\fontfamily{lmr}\selectfont,
}

\definecolor{findingcolor}{HTML}{9363bb}
\newtcolorbox[auto counter]{findings}{
  colback=gray!10,
  colframe=findingcolor!75,
  fonttitle=\bfseries,
title=Findings~\thetcbcounter.,
  breakable
}

\title{GameCraft-Bench: Can Agents Build Playable Games End-to-End in a Real Game Engine?}
\author{
Tongxu Luo$^{1,2*}$, Rongsheng Wang$^{1*}$, Jiaxi Bi$^{2,4*\dagger}$, Chenming Xu$^{1,2*}$, Zhengyang Tang$^{1,3*}$, Jianlong Chen$^{1}$, Juhao Liang$^{1}$, Ke Ji$^{1}$, Shuqi Guo$^{1,2}$, Yuhao Du$^{1,2}$, Fan Bu$^{1,2}$, Wenyu Du$^{5}$, Xiaotong Zhang$^{4}$, Kyle Li$^{7}$, Shaobo Wang$^{6}$, Linfeng Zhang$^{6}$, Yuxuan Liu$^{3}$, Xin Lai$^{3}$, Chenxin Li$^{3}$, Yiduo Guo$^{3}$, Zhexin Zhang$^{3}$, Xinyuan Wang$^{3}$, Tianyi Bai$^{3}$, Ziniu Li$^{3}$, Benyou Wang$^{1,2\ddagger}$\\
$^{1}$The Chinese University of Hong Kong, Shenzhen \quad
$^{2}$Shenzhen Loop Area Institute \\
$^{3}$Hunyuan Team, Tencent \quad
$^{4}$USTB \quad
$^{5}$DualverseAI \quad
$^{6}$SJTU \quad
$^{7}$NUS\\
$^{*}$Equal contribution \quad
$^{\dagger}$Work done during interning at SLAI \quad
$^{\ddagger}$Corresponding author
}

\abstract{
Game generation is an emerging application of coding agents, requiring models to transform natural-language specifications into playable interactive systems.
Unlike traditional coding tasks, game generation takes place within a game engine, where scripts, scenes, assets, rendering, and runtime interactions must jointly produce coherent gameplay.
We formalize end-to-end game generation as the problem of producing a complete game artifact that realizes a specification through observable player-game interaction in a target environment.
We argue that evaluating this setting requires three desiderata: \textbf{Engine Grounding}, \textbf{Artifact Completeness}, and \textbf{Interactive Verification}.
We propose an interaction-grounded evaluation framework that assesses executable gameplay through replayed demonstrations and rubric-guided multimodal judging.
We instantiate this framework as \benchmark{}, a benchmark comprising 140 Godot tasks across 15 game families.
Evaluations of frontier coding agents show that end-to-end game generation remains highly challenging: the strongest agent achieves only 41.46\%, and most agents score below 40\%.
Further analysis reveals that while agents often implement recognizable mechanics, they struggle to deliver complete games with sufficient content, functional visual feedback, and coherent presentation.
\textbf{See \url{https://tongxuluo.github.io/gamecraft-bench-website} for demos, code, and data.}
}
\begin{document}
\maketitle

\begin{figure}[ht]
  \centering
  \vspace{-12mm}
  \includegraphics[width=\linewidth]{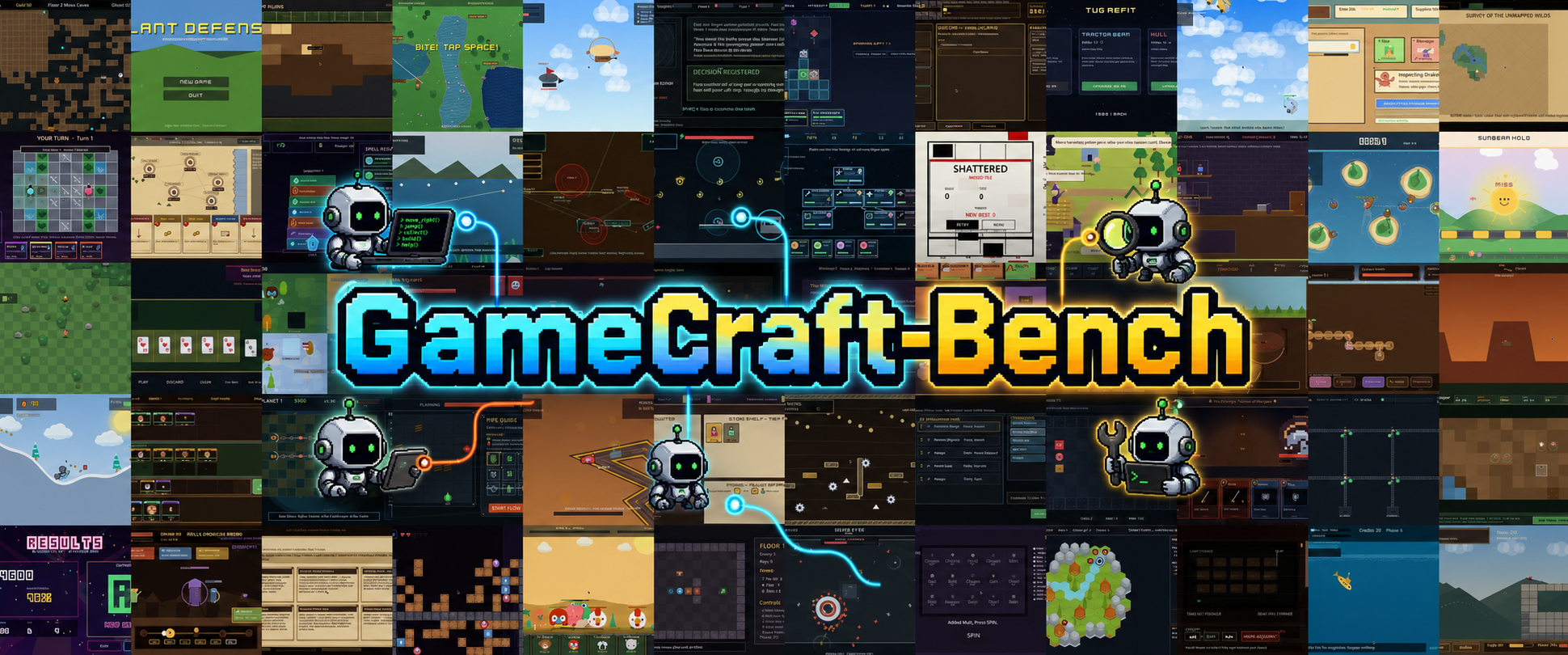}
  \caption{The playable games generated by coding agents in \benchmark, covering 15 families with 140 tasks.}
  \label{fig:teaser}
\end{figure}

\makeabstract

\begin{figure}[ht]
  \centering
  \includegraphics[width=\linewidth]{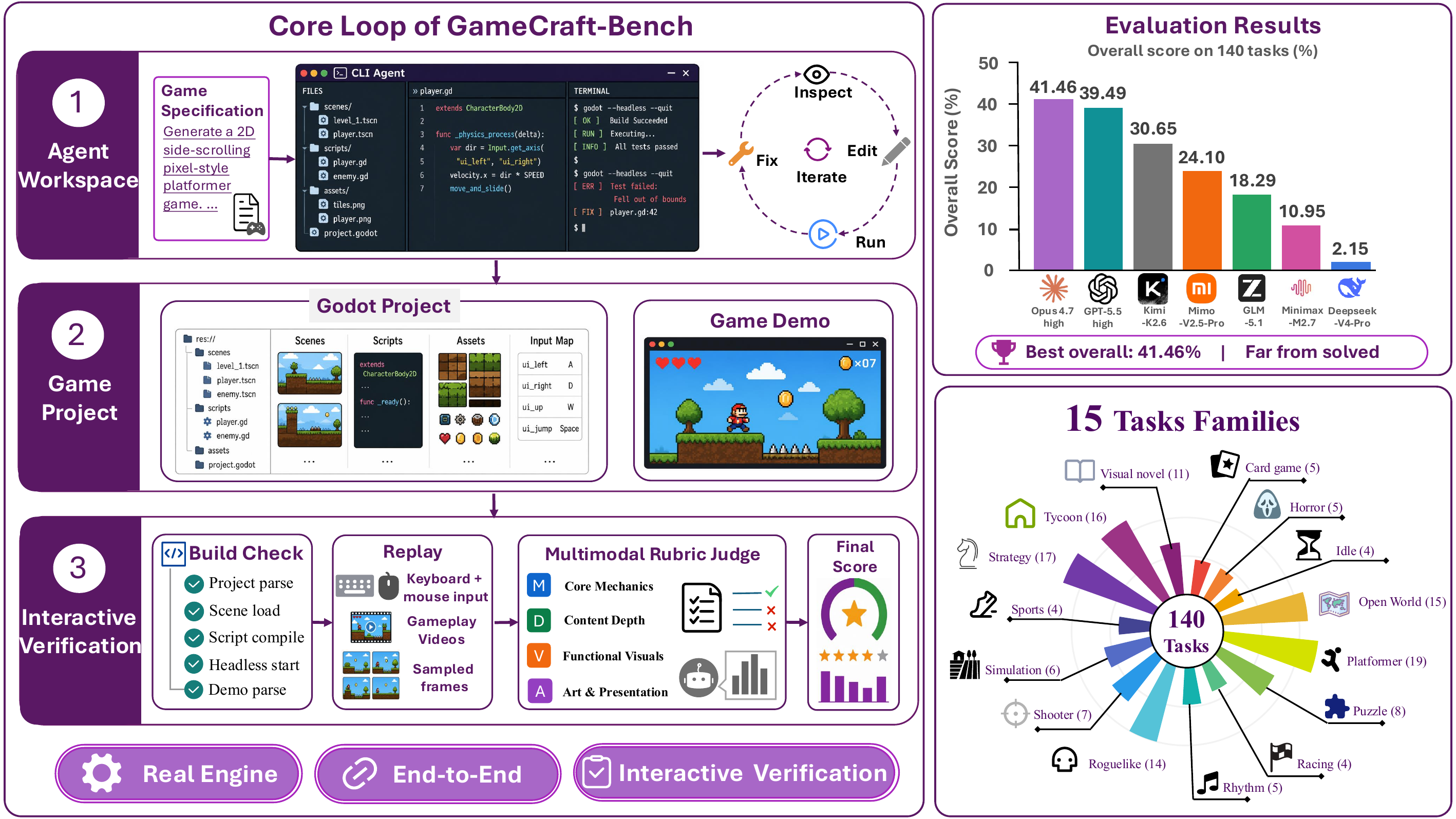}
  \caption{Overview of \benchmark{}. Agents turn natural-language game specifications into complete Godot projects, the verifier replays submitted interaction traces, and rubric-guided judging scores gameplay evidence.}
  \label{fig:overview}
\end{figure}

\section{Introduction}
Game generation~\cite{wan2025unigen,jiang2026opengame,ma2026creativegame,yin2026autoue} turns creative intent into interactive systems: players do not only view the result, but act inside it and expect the world to respond.
Recent coding agents~\cite{yang2024swe,kimiteam2026kimik25visualagentic,bolin2026unrolling, anthropic2026claudecode} make automated game generation increasingly plausible, because they can inspect files, edit projects, run tools, and iterate from execution feedback.
However, evaluating generated games is not the same as evaluating ordinary code: success depends on whether an artifact can be built, launched, played, and observed as an interactive system.

\paragraph{Desiderata for Game Generation.}
End-to-end game generation must answer three successive questions:
\emph{where} the agent develops,
\emph{what} artifacts it delivers,
and \emph{how} the resulting game is evaluated.
These questions motivate three desiderata:
\textbf{Desideratum~\ref{desideratum:1}: Engine Grounding}, which requires development in a real game engine with its native documentation and assets;
\textbf{Desideratum~\ref{desideratum:2}: Artifact Completeness}, which requires agents to produce a launchable game together with all necessary artifacts;
and \textbf{Desideratum~\ref{desideratum:3}: Interactive Verification}, which requires evaluation through direct interaction using player inputs and gameplay replays.
Together, these desiderata shift evaluation from \textbf{code correctness} to \textbf{interactive system correctness}, establishing a benchmark contract for engine-grounded, fully constructed, and behaviorally evaluated game generation.

\paragraph{The Existing Benchmarks Fail to Meet the Desiderata.}
Existing game generation benchmarks each satisfy part of this contract, but not all three desiderata.
OpenGame-Bench~\cite{jiang2026opengame} satisfies Artifact Completeness by asking agents to generate complete games from open-ended prompts; however, it targets web games rather than real-engine projects and does not judge games through gameplay interactions.
GameDevBench~\cite{chi2026gamedevbench} satisfies Engine Grounding by moving evaluation into real engines; however, it studies localized tutorial-derived edits and deterministic tests rather than complete game construction and interaction-based playability.
Closest to our setting, WebGameBench~\citep{zhang2026webgamebench} evaluates Web games through browser interaction, but it remains outside real game-engine development and relies on evaluation-side exploration.
Thus, existing benchmarks leave open the full setting captured by Table~\ref{tab:benchmark-comparison}: whether agents can carry user intent all the way to a complete engine-native game artifact whose behavior is judged through interaction.

\begin{table}[t]
  \centering
  \small
  \setlength{\tabcolsep}{5pt}
  \begin{tabular}{lccc}
    \toprule
    Benchmark
    & \makecell{Desideratum~\ref{desideratum:1}\\Engine Grounding}
    & \makecell{Desideratum~\ref{desideratum:2}\\Artifact Completeness}
    & \makecell{Desideratum~\ref{desideratum:3}\\Interactive Verification} \\
    \midrule
    GameDevBench~\cite{chi2026gamedevbench}
      & Godot & $\times$ (Single Operation) & $\times$ \\
    OpenGame-Bench~\cite{jiang2026opengame}
    & Web & $\checkmark$ & $\times$ \\
    WebGameBench$^*$~\cite{zhang2026webgamebench}
      & Web & $\checkmark$ & $\checkmark$ \\
    \rowcolor[HTML]{f4f0fa}
    \benchmark{}
      & Godot & $\checkmark$ & $\checkmark$ \\
    \bottomrule
  \end{tabular}
  \caption{Comparison with existing game generation benchmarks along the three desiderata. $^*$Notably, WebGameBench~\citep{zhang2026webgamebench} is concurrent with our work: it evaluates delivered Web games, while \benchmark{} targets complete projects in a dedicated game engine.}
  \label{tab:benchmark-comparison}
\end{table}

\paragraph{How \benchmark{} Fills the Gap.}
We introduce \benchmark{}, an end-to-end benchmark designed to jointly satisfy the three desiderata for game generation.
It grounds development in a real game engine (Godot), evaluates complete launchable game projects rather than partial artifacts, and verifies generated games through replayed interactions instead of static inspection alone.
The benchmark comprises 140 tasks across 15 game families, covering diverse gameplay requirements ranging from mechanics and progression systems to visual presentation.
By unifying \textbf{Engine Grounding}, \textbf{Artifact Completeness}, and \textbf{Interactive Verification}, \benchmark{} evaluates whether coding agents can transform natural-language game specifications into playable game experiences.

\paragraph{Observations on \benchmark.}
We evaluate frontier coding agents on \benchmark{} and find that end-to-end game generation remains far from solved.
Even the strongest agent achieves only 41.46\% overall, and most evaluated agents score substantially lower.
The low scores are not caused by a single missing capability: agents fail across core mechanics, content depth, functional visuals, and art and presentation, with performance varying sharply across game families.
These results suggest that current coding agents can often produce fragments of game-like software, but still struggle to reliably follow a game specification all the way to a playable, visually coherent artifact in a real engine.

\paragraph{Contributions.}
We summarize our contributions as follows:
\begin{itemize}[leftmargin=*]
\item We formalize end-to-end game generation as the problem of transforming game specifications into playable interactive systems, and identify three desiderata for its evaluation: \textbf{Engine Grounding}, \textbf{Artifact Completeness}, and \textbf{Interactive Verification}.
\item We propose an interaction-grounded evaluation framework that verifies generated games through executable gameplay evidence, combining replay-based interaction protocols with rubric-guided assessment of mechanics, content, visual functionality, and presentation quality.
\item We instantiate this framework as \benchmark{}, a benchmark comprising 140 tasks across 15 game families in Godot, where agents must generate complete game projects from natural-language game specifications.
\item We benchmark frontier coding agents and provide diagnostic analyses showing that current systems remain far from reliable end-to-end game generation, with limitations spanning mechanics, content depth, visual feedback, presentation quality, and task-completion behavior.
\end{itemize}

\section{What Should Be a Good Game Generation Benchmark?}
\subsection{Problem Definition}
\begin{figure}[ht]
\centering
\includegraphics[width=0.7\linewidth]{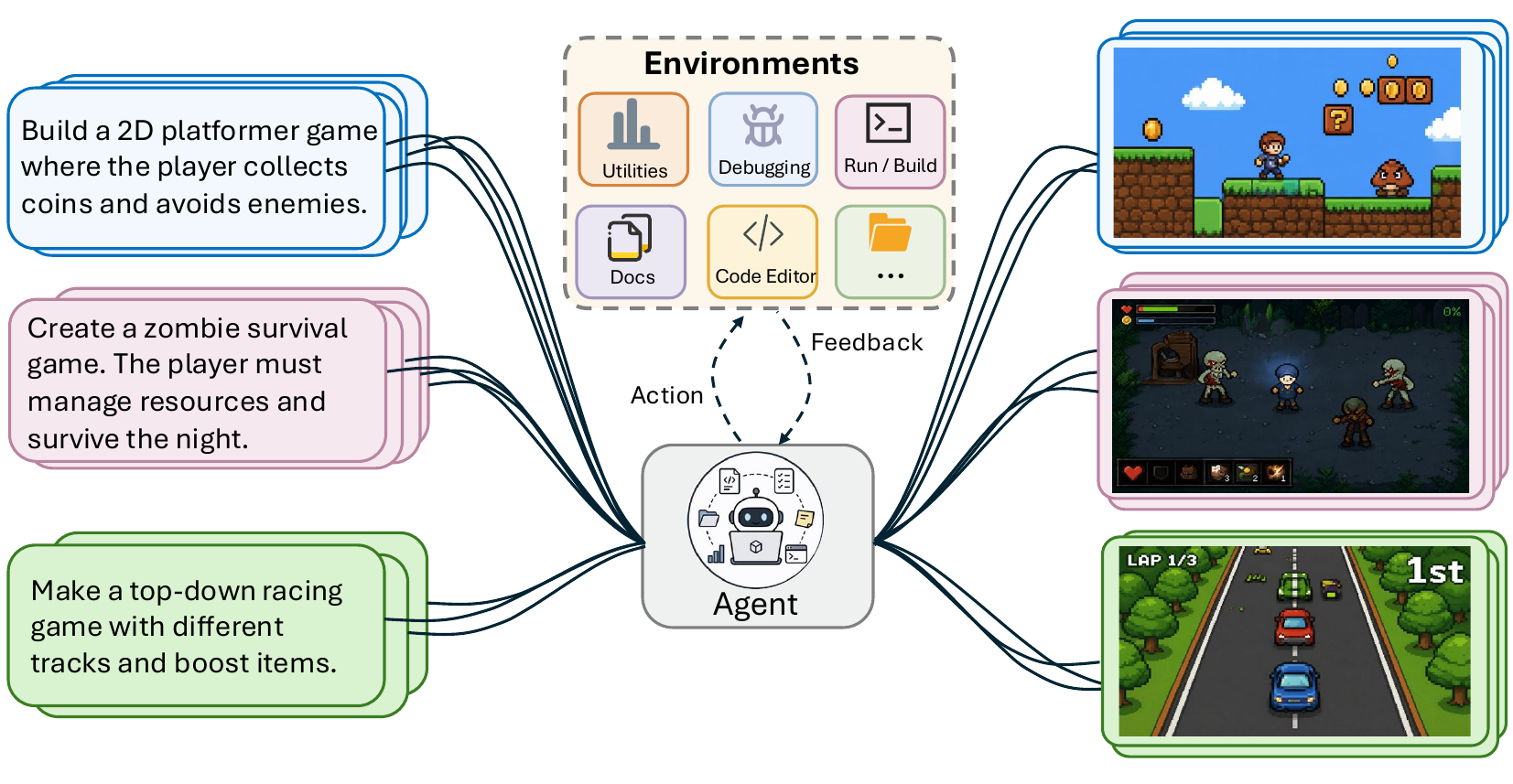}
\caption{Problem definition for AI game generation. An agent transforms natural-language game specifications \(s\) into playable game artifacts \(G\) by acting
within a development and runtime environment \(\mathcal{E}\).}
\label{fig:problem-definition}
\end{figure}

AI game generation studies \textit{how an agent turns a high-level design intent into a playable interactive system}.
As illustrated in Figure~\ref{fig:problem-definition}, an agent receives a game specification, operates within a development and runtime environment, and produces a playable game artifact.
We abstract this process as
\[
x=(s,\mathcal{E}) \quad \longmapsto \quad y=G,
\]
where \(s\) is a game specification describing the intended player experience, including rules, mechanics, goals, content, and presentation; \(\mathcal{E}\) is the target development and runtime environment; and \(G\) is the generated game artifact.
A successful output \(G\) should be launchable in \(\mathcal{E}\) and should realize \(s\) through observable responses to player actions.

\subsection{Three Desiderata.}
\begin{figure}[ht]
  \centering
    \includegraphics[width=0.7\linewidth]{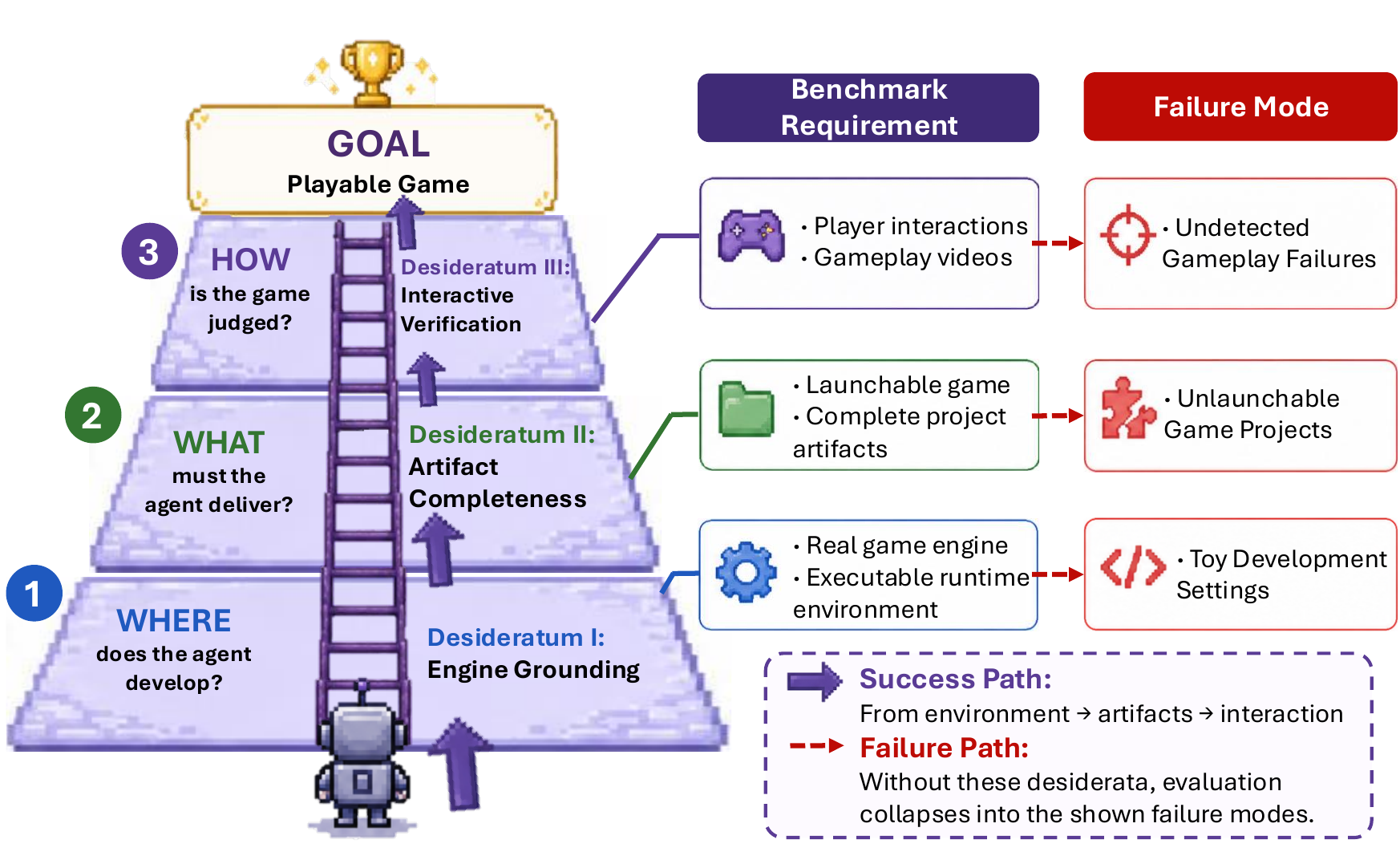}
  \caption{Three desiderata for evaluating end-to-end game generation.}
  \label{fig:desiderata}
\end{figure}

A good benchmark must preserve the full meaning of the mapping \((s,\mathcal{E})\mapsto G\).
This requires constraining three spaces of the evaluation problem: the \textbf{development space}, or \textit{where} the agent constructs the game; the \textbf{output space}, or \textit{what} artifact the agent must deliver; and the \textbf{evaluation space}, or \textit{how} the delivered artifact is judged.
For game generation, these spaces induce three desiderata: \textbf{Engine Grounding}, \textbf{Artifact Completeness}, and \textbf{Interactive Verification}.
As illustrated in Figure~\ref{fig:desiderata}, the desiderata are jointly complete because they cover the full success path of game generation: generation inside an environment, delivery of a game artifact, and verification through interaction.
If any one space is left unconstrained, the benchmark no longer evaluates end-to-end game generation: it may collapse into toy development settings, partial artifacts, or static assessments that miss gameplay failures.
Together, the three desiderata shift evaluation from code correctness or visual plausibility to interactive system correctness.

\begin{desideratum}\label{desideratum:1}
\textbf{Engine Grounding}. Games should be evaluated within a \textbf{concrete} game generation and runtime \textbf{environment}. 
\end{desideratum}
Game behavior is not defined by source code alone, but by engine-level semantics such as scene hierarchy, scripting lifecycle, asset loading, input dispatch, physics, rendering, project configuration, and launch procedure.
Engine Grounding therefore requires the benchmark to preserve these runtime constraints, so the agent must solve the integration problems of real game generation rather than an abstract programming exercise that only imitates game logic.
Without it, evaluation can collapse into toy program synthesis: locally plausible code may be produced without demonstrating that it works inside the operational environment where games actually run.

\begin{desideratum}\label{desideratum:2}
\textbf{Artifact Completeness}. Games should be evaluated as a \textbf{complete} launchable \textbf{artifact}.
\end{desideratum}
Playability emerges only when the required components are assembled into a functioning whole: project metadata, entry scenes, scripts, assets, UI elements, input mappings, configuration files, and runtime resources must all be present and connected.
Artifact Completeness therefore makes the launchable game project the unit of evaluation, rather than isolated scripts, sprites, mechanics prototypes, or visually plausible scenes that require further human assembly.
This does not require a polished commercial game, but the artifact must be self-contained enough for the intended game loop to run in the target environment.
Without this desideratum, a benchmark can over-credit partial work that looks game-like in isolation but fails at the system boundary where a game must become a runnable experience.

\begin{desideratum}\label{desideratum:3}
\textbf{Interactive Verification}. Games should ultimately be judged by what happens when they are \textbf{played interactively}.
\end{desideratum}
The defining property of a game is the action--response loop: the player acts, the system updates, and the world presents consequences that sustain goals, feedback, challenge, progression, or narrative.
Many important failures are invisible before this loop is exercised, including unresponsive controls, incorrect collisions, inactive enemies, unreachable objectives, missing UI feedback, broken state transitions, or visual scenes that become illegible during play.
Interactive Verification therefore requires the benchmark to judge observed gameplay behavior rather than static inspection alone: the artifact must be run, acted upon, and observed as an interactive system, whether through human play, scripted inputs, replayed demonstrations, learned policies, or another standardized interaction procedure.
Without this desideratum, evaluation may certify artifacts that look correct before play while failing precisely when a game must respond to the player.

\section{The \benchmark{}}
\subsection{Task Definition}
\benchmark{} instantiates the general game-generation mapping \((s,\mathcal{E})\mapsto G\) in a real game-engine setting.
Each task is defined as 
\[
\tau = (s,\mathcal{E},\rho),
\]
where \(s\) is the game specification given to the agent, \(\mathcal{E}\) is the target development and runtime environment, and \(\rho\) is the hidden rubric used by the verifier.
In \benchmark{}, the environment \(\mathcal{E}\) is operationalized as
\[
\mathcal{E}=(\mathcal{R},\mathcal{W},\mathcal{A},\mathcal{C}),
\]
where \(\mathcal{R}\) is the Godot engine runtime and toolchain, \(\mathcal{W}\) is the editable workspace, \(\mathcal{A}\) is the shared resource interface, and \(\mathcal{C}\) is the submission contract.
The agent observes \((s,\mathcal{E})\), but not \(\rho\).

Given \((s,\mathcal{E})\), the agent must produce
\[
    y=(G,\Pi),
\]
where \(G\) is a complete Godot project and \(\Pi=\{\pi_i\}_{i=1}^{n}\) is a set of replayable interaction traces.
Notably, the traces are not part of the game artifact itself; they provide standardized interaction evidence for evaluation.
The verifier launches \(G\) in \(\mathcal{R}\) and replays \(\Pi\) to obtain gameplay observations:
\[
O=\mathrm{Replay}_{\mathcal{R}}(G,\Pi).
\]
A successful submission is one whose observed behavior \(O\) realizes the specification \(s\) under the constraints of \(\mathcal{E}\).

\begin{figure}[ht]
\centering
\includegraphics[width=\linewidth]{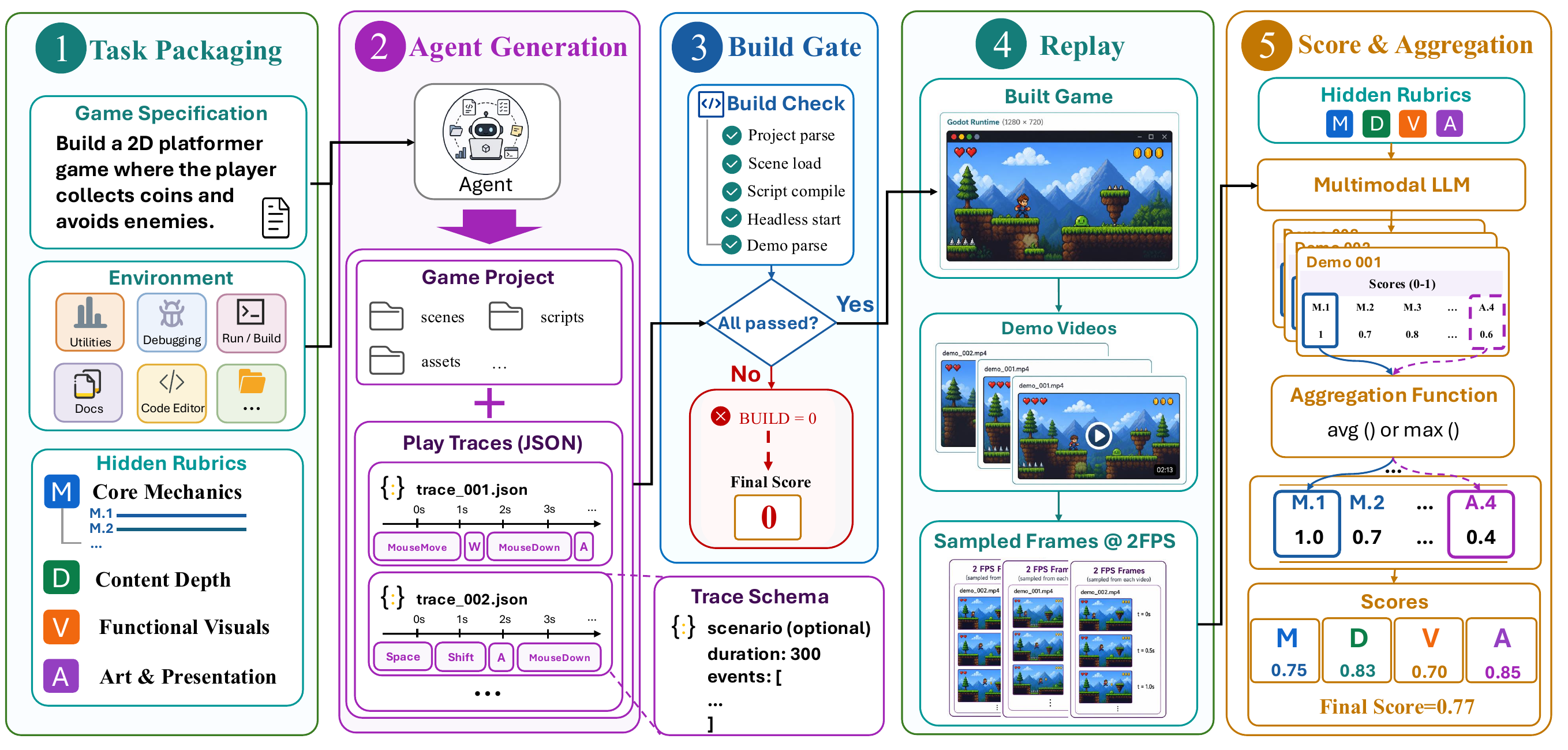}                      
\caption{End-to-end evaluation pipeline of \benchmark{}.
A task packages a game specification, Godot environment, and hidden rubric; the agent produces a complete game project and replay traces; the verifier checks launchability, replays traces into gameplay videos and sampled frames, and scores the resulting evidence with rubric-guided multimodal judging.}
\label{fig:verify-pipeline}
\end{figure}

\subsection{Implementation}
\benchmark{} implements each task as the five-stage pipeline shown in Figure~\ref{fig:verify-pipeline}.
The pipeline follows the full lifecycle of an evaluation instance: a task is packaged, an agent builds and submits a game, the verifier checks whether the submission is executable, replay traces are converted into gameplay evidence, and the evidence is finally scored against a hidden rubric.
This structure makes open-ended game generation comparable without forcing all games to share the same controls, levels, objectives, or termination conditions.

\paragraph{Stage 1: Task Packaging.}
Each task packages three objects: a natural-language game specification, a Godot-based development environment, and a hidden evaluation rubric.
The specification describes the intended player experience, including the player fantasy, core loop, mechanics, goals, content scope, visual style, and expected demonstration scenarios.
The environment provides the editable workspace, Godot runtime and toolchain, shared assets, documentation, helper tools, and the submission contract.
The hidden rubric decomposes the same intent into observable requirements under four categories: Core Mechanics, Content Depth, Functional Visuals, and Art and Presentation.
Thus, the agent receives a natural game-building specification, while the verifier retains a precise scoring target that is not exposed as a checklist.

\paragraph{Stage 2: Agent Generation.}
The agent then uses the packaged environment to construct a Godot project.
It may create scenes, write scripts, configure input actions, use shared assets, run the project, inspect screenshots, and iterate from tool feedback.
The required submission has two parts: a complete Godot game project \(G\) and a set of replayable demonstration traces \(\Pi\).
The project is the generated artifact to be evaluated.
The traces are evaluation artifacts: they contain timed mouse and keyboard events, optionally with scenario identifiers, that specify how the verifier should exercise the game to reveal relevant gameplay behavior.

\paragraph{Stage 3: Build Gate.} 
The verifier first checks whether the submitted project and traces can enter the evaluation pipeline.
It parses the project structure and trace files, launches the Godot project under the benchmark runtime, and checks whether the game reaches a runnable state.
If the project fails to launch, or if no submitted trace can be parsed, the verifier sets \(\mathrm{BUILD}=0\) and the final score is zero.
This gate ensures that scoring begins only after the submission becomes an executable artifact with replayable interaction evidence.

\paragraph{Stage 4: Replay.}
For each valid trace, the verifier starts a fresh Godot instance and replays the recorded mouse and keyboard events in a fixed \(1280\times720\) viewport.
Replay is the bridge between open-ended interaction and standardized evidence.
Without submitted traces, the verifier would first have to infer how each game should be played before judging whether it works, turning verification into an autonomous exploration problem and making scores depend on the verifier's exploration policy.
By replaying demonstrations, \benchmark{} keeps the interaction procedure fixed while still evaluating the artifact under actual player input.
The verifier records one gameplay video per replay and samples frames at two frames per second.
The resulting videos and frames are the evidence passed to the judge.

\paragraph{Stage 5: Scoring and Aggregation.}
The verifier scores the replayed gameplay evidence with the hidden task rubric.
The rubric categories are grounded in playable-artifact principles~\citep{hunicke2004mda,desurvire2004using}: following MDA~\citep{hunicke2004mda}, \textbf{Core Mechanics} captures rules and state transitions, \textbf{Content Depth} captures runtime scope and progression, and \textbf{Art and Presentation} captures the authored sensory layer; following playability heuristics~\citep{desurvire2004using}, \textbf{Functional Visuals} captures readability and visual feedback during gameplay.
Each demo is scored independently by a multimodal judge using the sampled frames, evaluation prompt, and rubric items.
The verifier then aggregates demo-level scores according to item semantics: scenario-specific requirements use maximum aggregation, while persistent requirements use mean aggregation.
The aggregated item scores produce category scores \(\mathrm{M}\), \(\mathrm{D}\), \(\mathrm{V}\), and \(\mathrm{A}\), which are combined as
\[
\mathrm{Score}=\mathrm{BUILD}\times (w_M\mathrm{M}+w_D\mathrm{D}+w_V\mathrm{V}+w_A\mathrm{A}),
\]
where the four categories denote Core Mechanics, Content Depth, Functional Visuals, and Art and Presentation, with default weights \(0.15,0.35,0.15,0.35\), respectively.
We prioritize Content Depth and Art and Presentation because a complete game should extend beyond minimal mechanics and functional prototypes.

\subsection{How It Fulfills the Three Desiderata}
\paragraph{Grounded on a Real Engine: Godot (Desideratum~\ref{desideratum:1})}
\begin{table}[h]
\centering
\small
\setlength{\tabcolsep}{4pt}
\begin{tabular}{lcccc}
  \toprule
  \textbf{Benchmark Requirement}
  & \textbf{Godot} & \textbf{Unity} & \textbf{Unreal} & \textbf{GameMaker} \\
  \midrule
  Open-source
    & $\checkmark$ & $\times$ & $\times$ & $\times$ \\
  Lightweight setup
    & $\checkmark$ & $\triangle$ & $\times$ & $\checkmark$ \\
  Command-line/headless-friendly
    & $\checkmark$ & $\triangle$ & $\triangle$ & $\triangle$ \\
  Native 2D workflow
    & $\checkmark$ & $\triangle$ & $\triangle$ & $\checkmark$ \\
  Text-based scene/files
    & $\checkmark$ & $\triangle$ & $\triangle$ & $\times$ \\
  \bottomrule
\end{tabular}
\caption{Engine fit for automated 2D game-generation benchmarking. $\triangle$
indicates support with heavier setup, editor/toolchain constraints, or weaker fit for
lightweight reproducible evaluation.}
\label{tab:engine-comparison}
\end{table}

\benchmark{} uses Godot 4 as the concrete engine environment.
The goal is not to claim that Godot is the only automatable game engine, but to choose an engine that makes large-scale, reproducible, execution-based evaluation practical.
Godot is open source, lightweight, provides a native 2D engine stack, stores scenes in text-based project files, and supports command-line/headless execution for running and exporting projects.
These properties let the verifier launch generated games, replay traces, record gameplay evidence, and inspect failures without relying on a GUI editor workflow.
By contrast, Unity, Unreal, and GameMaker also support automation in different forms, but they introduce heavier installations, proprietary licensing constraints, editor-centric workflows, binary project formats, or weaker fit for lightweight 2D benchmark execution.
For this reason, \benchmark{} uses Godot as the benchmark environment while leaving multi-engine evaluation to future work.

\paragraph{Full Game Delivery (Desideratum~\ref{desideratum:2})}
\benchmark{} makes the generated game artifact \(G\) the unit of submission.
The submission contract \(\mathcal{C}\) requires agents to place a complete Godot project under the target workspace, including project metadata, entry scenes, scripts, copied assets, input configuration, UI resources, and the runtime entry point needed to launch the game.
This prevents agents from receiving credit for isolated mechanics, loose assets, partial scenes, or code fragments that require additional human assembly.
The verifier enforces this requirement with a launchability gate: if the project cannot be launched in the Godot runtime, \(\textsc{Build}=0\) and the final score is zero.
Thus, \benchmark{} evaluates whether an agent can deliver a self-contained playable artifact, not merely produce game-like components.

\paragraph{Interactive Evaluation (Desideratum~\ref{desideratum:3})}
\benchmark{} grounds evaluation in observed player-game interaction rather than static inspection.
Each submission includes replayable traces \(\Pi\), and the verifier executes the submitted project \(G\) in the Godot runtime, replays the traces, and records gameplay observations \(O=\mathrm{Replay}_{\mathcal{R}}(G,\Pi)\).
The hidden rubric \(\rho\) is then applied to this gameplay evidence, scoring whether the requested mechanics, content, visual feedback, and presentation actually appear during play.
This ensures that artifacts are evaluated as interactive systems: source code, project structure, assets, and screenshots may be useful for debugging, but they do not substitute for observed behavior under player input.

\subsection{Task Suite and Annotation Quality}
\benchmark{} contains 140 tasks across 15 game families.
The families are chosen to cover distinct game-construction demands rather than surface genres alone: continuous control and collision in platformers and racing games, rule and state management in card and strategy games, progression and economy in tycoon and idle games, exploration and spatial layout in open-world and roguelike games, and narrative or presentation-heavy interaction in visual novels, horror, rhythm, and sports games.
Table~\ref{tab:families} summarizes the family-level coverage.

\paragraph{Annotation Process.}
We use Harbor~\cite{Harbor_Framework} as the task-authoring framework.
The task suite is annotated by 12 annotators, each selected for extensive gameplay experience across diverse genres.
For each task, an annotator authors two aligned artifacts: a public product-style game specification that describes the player fantasy, core loop, mechanics, and art direction; and a hidden evaluation rubric that decomposes the same game into observable, requirement-level criteria.
The annotator ensures that the specification is natural and open-ended—specifying the intended experience without prescribing implementation details—while the rubric is precise enough to be scored from replayed gameplay evidence.
The instruction given to annotators is summarized below.

\begin{prompt}{Instruction for Annotators}
Annotate each GameCraft-Bench task into three files: \texttt{task.toml},
\texttt{instruction.md}, and \texttt{tests/rubric.json}. The final goal is to
evaluate whether an agent can build a complete, shippable micro-game, not a
prototype, static UI mockup, or label-only demo.

\textbf{1. Metadata.} Use \texttt{task.toml} only for metadata and runtime
configuration, such as the task name, short description, keywords, verifier
settings, agent budget, and environment settings. Do not put gameplay
requirements, demo rules, or rubric items in \texttt{task.toml}.

\textbf{2. Game specification.} Use \texttt{instruction.md} as the source of truth for
what the agent must build. The opening should name the game, specify the output
path \texttt{/workspace/game}, and set the quality target as a complete,
shippable micro-game. Write a \emph{Core Vision} section describing the player
fantasy and core loop, and a \emph{What the Player Experiences} section
describing the player-facing arc. The specification should describe observable player
experience, meaningful decisions, feedback, variation, and outcomes, while
avoiding rubric-style checklists or implementation-level commands. Include
standard sections for assets, project layout, launch command, demos, scenarios,
and trace format. Do not ask for audio requirements, because the verifier
scores visual evidence.

\textbf{3. Evaluation rubric.} Use \texttt{tests/rubric.json} to translate the
game specification into concrete rubric items for the verifier. Each rubric should
include \texttt{score\_formula}, \texttt{max\_demos},
\texttt{max\_demo\_seconds}, \texttt{build\_check}, \texttt{categories}, and
\texttt{requirements}. Use four fixed categories: Core Mechanics, Content
Depth, Functional Visuals, and Art \& Presentation. Use stable item prefixes:
\texttt{M*}, \texttt{D*}, \texttt{V*}, and \texttt{A*}. Keep the total number
of rubric items at 24 or fewer.

\textbf{4. Category calibration.} Core Mechanics should evaluate whether the
irreducible gameplay loop works through player action, state change, rules,
consequences, goals, and failure. Content Depth should evaluate varied material,
progression, late-game states, or multiple scenarios. Functional Visuals should
evaluate readability and feedback during play, including readable state,
warnings, objectives, and stable UI at 1280$\times$720. Presentation \& Art
should evaluate whether the game feels authored and publishable rather than
debug-like.

\textbf{5. Scoring calibration and aggregation.} Full credit should require
publishable vertical-slice quality. Cap weak evidence at 0.5 or below when a
feature is label-only, decorative, static, disconnected from gameplay,
represented only by names or colors, dominated by default widgets, or built
primarily from raw programmatic shapes. Use \texttt{mean} for properties that
should hold across demos, such as readability, visual feedback, art
consistency, UI overlap, and presentation quality. Use \texttt{max} only when
observing a feature once is enough to prove it exists, and use it
conservatively to avoid over-crediting static demonstrations.
\end{prompt}

\paragraph{Quality Control.}
After the specification and rubric are drafted, annotators enter a validation phase by writing a simple oracle solution—a minimal playable sketch in Godot—to cross-check the task's internal consistency.
The oracle serves three validation functions: it verifies whether the specification is implementable in the engine, whether the requested behavior can be demonstrated through replayed gameplay, and whether every rubric item corresponds to an observable state rather than a subjective preference.
If the oracle cannot reach the intended gameplay state, or if a rubric item lacks supporting evidence in the replay, the annotator revises the specification and rubric and re-validates until the task becomes internally consistent.
This specification--rubric--oracle alignment ensures that the agent faces a well-defined target, the judge evaluates only executable behavior, and the final score reflects whether the generated artifact realizes the intended player experience.

\begin{table}[ht]
  \centering
  \small
  \begin{tabular}{lr|lr|lr}
    \toprule
    Family & Tasks & Family & Tasks & Family & Tasks \\
    \midrule
    Platformer & 19 & Strategy & 17 & Tycoon & 16 \\
    Open-world & 15 & Roguelike & 14 & Visual novel & 11 \\
    Puzzle & 8 & Shooter & 7 & Simulation & 6 \\
    Card game & 5 & Horror & 5 & Rhythm & 5 \\
    Idle & 4 & Racing & 4 & Sports & 4 \\
    \bottomrule
  \end{tabular}
  \caption{Task family coverage in \benchmark{}. Counts exclude the example task.}
  \label{tab:families}
\end{table}

\section{Benchmarking Results}
\paragraph{Experimental Setup.}
We evaluate seven frontier coding-agent configurations on the full \benchmark{} task suite: Claude Code~\cite{anthropic2026claudecode} with Opus-4.7~\cite{anthropic2026} high and MiMo-V2.5-Pro~\cite{mimo2026v25pro}, Codex~\cite{bolin2026unrolling} with GPT-5.5~\cite{openai2026gpt55} high and DeepSeek-V4-Pro~\cite{deepseekai2026deepseekv4}, Kimi Code with Kimi-K2.6~\cite{kimiteam2026kimik25visualagentic}, and Code Buddy with GLM-5.1~\cite{glm5team2026glm5vibecodingagentic} and MiniMax-M2.7~\cite{minimax2026m27}.
Each configuration is run on all 140 tasks under the same benchmark interface and environment.
We provide detailed evaluation setup in Appendix~\ref{app:evaluation-details}.

\begin{figure}[ht]
    \centering
    \includegraphics[width=\linewidth]{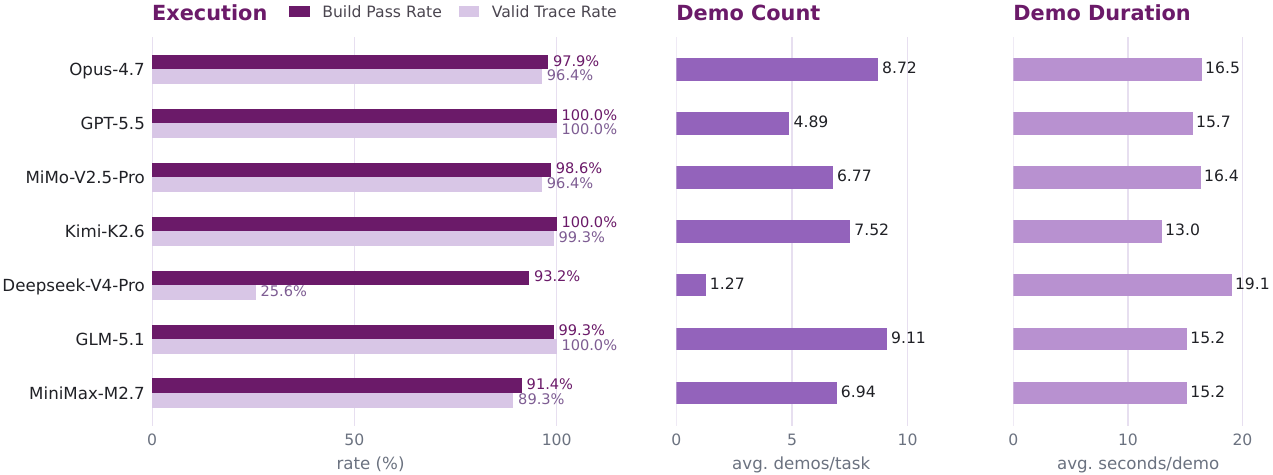}
    \caption{Execution and replay statistics across agents. Build pass measures launchability, valid trace measures whether at least one submitted demonstration can be replayed, and demo count/duration summarize the amount of gameplay evidence available for judging.}
\label{fig:runtime-stats}
\end{figure}

\begin{table}[ht]     
\centering            
\begin{tabular}{llccccc}
\toprule
\textbf{Harness} & \textbf{Model} & \textbf{Overall} & \textbf{Mechanics} &   
\textbf{Depth} & \textbf{Visuals} & \textbf{Art} \\
\midrule
\multirow{2}{*}{Claude Code}
& \modellogo{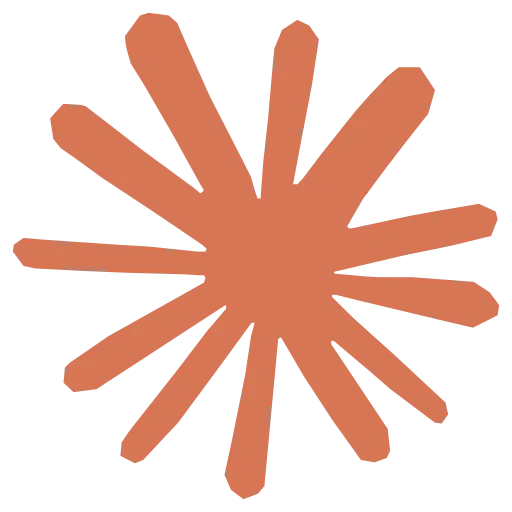}\,Opus-4.7 high
& \textbf{41.46} & \textbf{55.34} & \textbf{39.48}
& \textbf{42.78} & \textbf{36.86} \\
& \modellogo{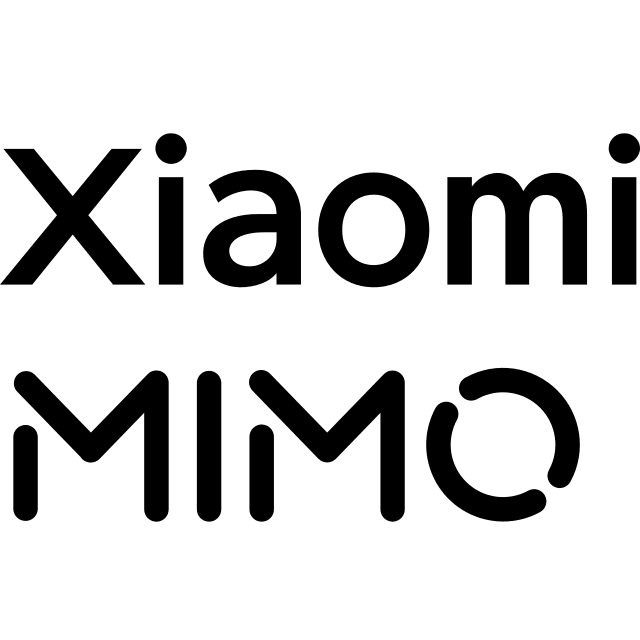}\,MiMo-V2.5-Pro
& 24.10 & 32.33 & 22.59 & 27.45 & 20.65 \\
\midrule
\multirow{2}{*}{Codex}
& \modellogo{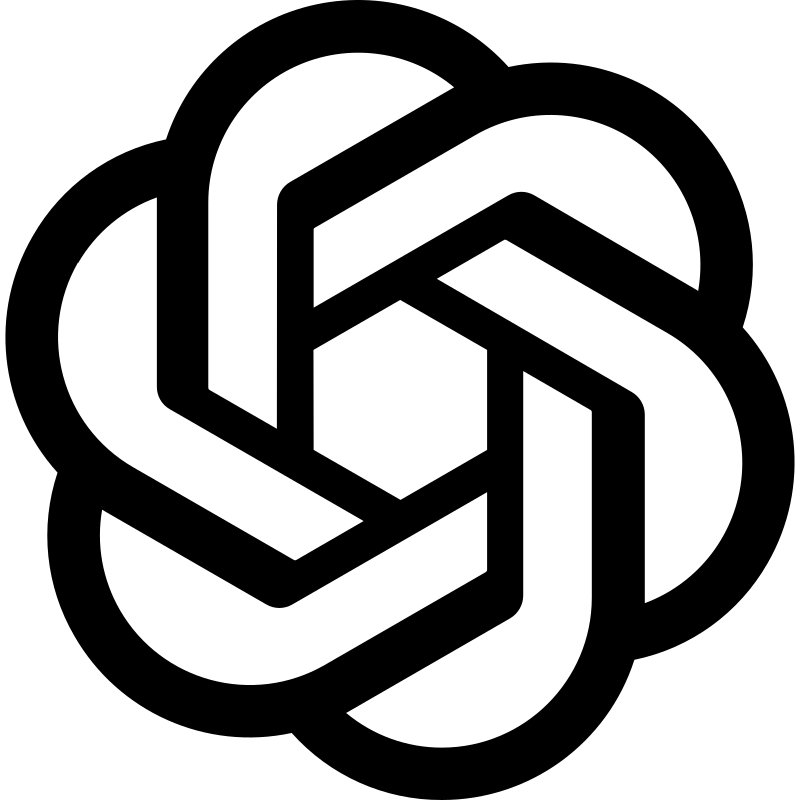}\,GPT-5.5 high
& \underline{39.49} & \underline{54.36} & \underline{38.61}
& \underline{41.84} & \underline{32.94} \\
& \modellogo{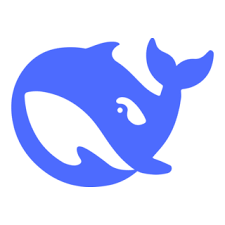}\,DeepSeek-V4-Pro
& 2.15 & 2.25 & 1.69 & 1.97 & 2.63 \\
\midrule
\multirow{1}{*}{Kimi Code}
& \modellogo{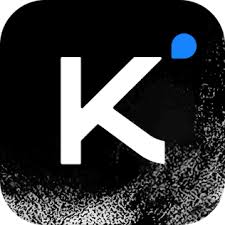}\,Kimi-K2.6
& 30.65 & 39.76 & 28.07 & 33.66 & 27.99 \\
\midrule
\multirow{2}{*}{Code Buddy}
& \modellogo{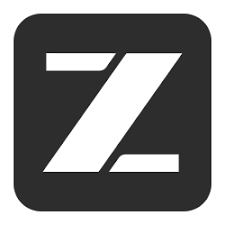}\,GLM-5.1
& 18.29 & 25.23 & 17.80 & 21.14 & 14.59 \\
& \modellogo{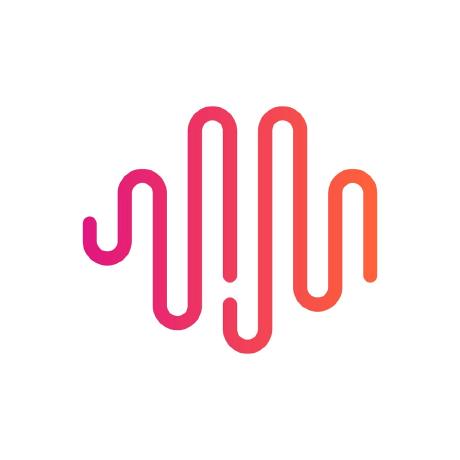}\,MiniMax-M2.7
& 10.95 & 14.27 & 9.92 & 14.92 & 8.85 \\
\bottomrule
\end{tabular}
\caption{Benchmark-level results (\%). Mechanics, Depth, Visuals, and Art correspond to the four rubric categories. Best and second-best scores are bolded and underlined.}
\label{tab:main-results}
\end{table}

\paragraph{Main Results.} Table~\ref{tab:main-results} reports benchmark-level scores in percentage points; detailed family-level results are provided in Appendix~\ref{app:family-results}.
The strongest completed configuration, Opus-4.7 high under Claude Code, reaches only 41.46\% overall.
GPT-5.5 high follows closely at 39.49\%, while Kimi-K2.6 reaches 30.65\%.     
MiMo-V2.5-Pro drops to 24.10\%, and DeepSeek-V4-Pro reaches only 2.15\%.      
Thus, even the best current coding agents remain far below reliable end-to-end game generation: they can often produce runnable game-like artifacts, but they do not consistently realize the requested mechanics, content, visual state, and presentation.
Figure~\ref{fig:runtime-stats} further separates launchability from replayable evidence.
Most strong agents satisfy the basic execution contract, but low final scores show that producing a runnable project and replayable traces is still far from satisfying the rubric.
DeepSeek-V4-Pro often ignores or violates the demonstration-trace requirement, producing a large gap between build pass rate and valid-trace rate.

\paragraph{Category-level Results.}
The benchmark-level category scores show that mechanics are consistently stronger than content and presentation.
Among the stronger agents, Opus-4.7 high, GPT-5.5 high, and Kimi-K2.6 obtain 55.34\%, 54.36\%, and 39.76\% on Core Mechanics.
The same agents score lower on Content Depth, at 39.48\%, 38.61\%, and 28.07\%, and lower still on Art and Presentation for Kimi-K2.6 and MiMo-V2.5-Pro.
This pattern suggests that agents can often create partial interaction loops, but struggle to expand them into complete games with enough state progression, visual readability, and polished presentation.

\begin{findings}
Agents more often produce recognizable local mechanics, but still fail to assemble them into complete, coherent interactive systems.
\end{findings}

\section{In-depth Analysis}
\subsection{On the Diagnostic Patterns of Agents}
\begin{figure}[ht]
  \centering
  \includegraphics[width=\linewidth]{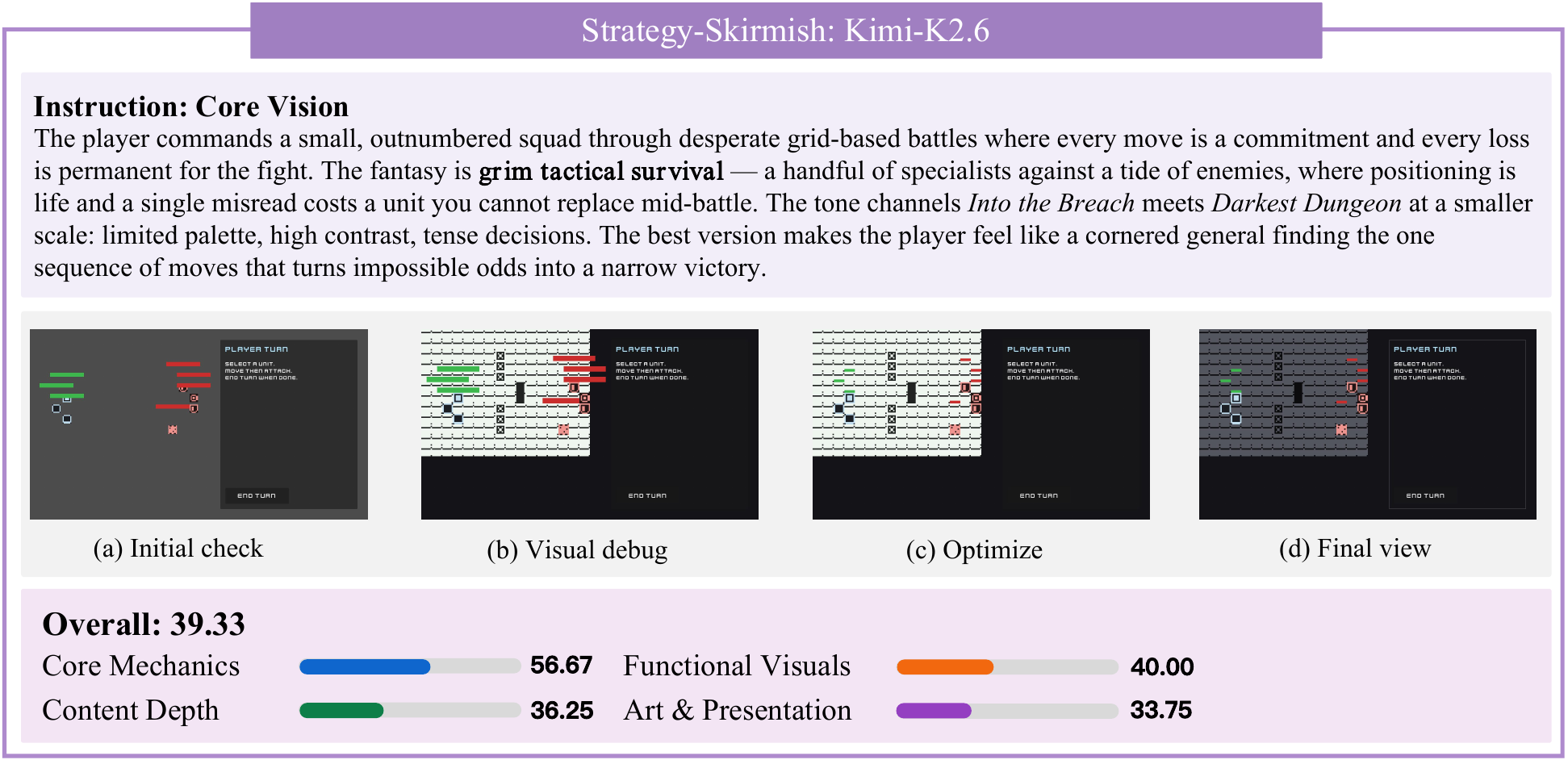}
  \caption{Example of perception-guided debugging in Kimi-K2.6.
Repeated inspection of rendered gameplay enables the agent to identify and correct player-facing failures.}
  \label{fig:visual-iteration-case}
\end{figure}

\paragraph{Success Signal: Do Agents Use Rendered Feedback?}
Game generation is difficult to debug from source code and terminal logs alone.
Many failures only become visible after rendering the project: incorrect camera framing, unreadable UI, missing visual feedback, broken level layout, or demo states that fail to reveal the intended mechanic.
Without visual feedback, an agent may continue editing code while remaining blind to whether the player-facing game state is actually coherent.
Visual interaction changes this loop by turning rendered gameplay into a debugging signal: the agent can inspect the screen, locate mismatches between the intended and observed game state, and revise the project accordingly.
Kimi-K2.6 provides a concrete example of this behavior.
Across the 140 Kimi tasks, we count 2,998 rendered-screen inspections, averaging 21.41 per task with a median of 19; only 4 tasks use no screenshots.
The same pattern appears in Opus-4.7, which invokes the helper 1,952 times (13.94 per task), whereas GPT-5.5 uses it much more sparingly, with 268 calls (1.91 per task).
Figure~\ref{fig:visual-iteration-case} shows a representative case from \textit{Strategy-Skirmish}, where Kimi repeatedly inspected rendered frames, debugged misaligned unit placement and missing selection highlights, and adjusted the grid and turn-indicator layout.
This suggests that visual interaction turns game generation from one-shot code synthesis into perception-guided iteration.

\begin{findings}
Rendered interaction helps agents debug failures that are not observable in source code.
\end{findings}

\begin{figure}[ht]
  \centering
  \includegraphics[width=0.90\linewidth]{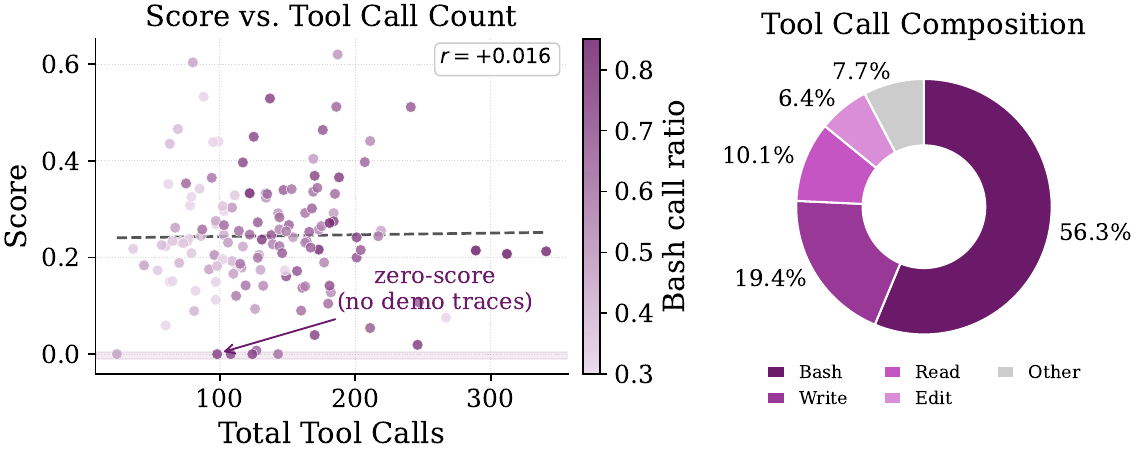}
  \caption{Tool usage of MiMo-V2.5-Pro across 140 tasks.
        \textbf{\textit{Left}:} Score vs.\ total tool calls, colored by per-task
        \texttt{Bash} call fraction. The near-zero correlation
        ($r = {+}0.016$) indicates that tool usage volume is decoupled
        from task quality.
        \textbf{\textit{Right}:} Aggregate tool call composition; shell execution
        dominates at 56.3\%, while code-reading and editing account for
        only 16.5\%.}
  \label{fig:mimo-reward-tools}
\end{figure}

\paragraph{Failure Signal: Do More Tools Help?}
Writing more code or running more commands is not sufficient for producing a playable game.
MiMo-V2.5-Pro exhibits a characteristic \textit{write-first, bash-heavy} pattern: the agent rapidly scaffolds a complete file set, including \texttt{project.godot}, GDScript sources, and scene files, before any validation.
It then enters an extended execution-driven debug phase dominated by shell commands, which account for 56.3\% of all tool calls across 140 tasks, while code-reading and editing operations (\texttt{Read} + \texttt{Edit}) account for only 16.5\%.
Crucially, total tool usage is decoupled from task quality, as shown in Figure~\ref{fig:mimo-reward-tools}: a typical task involves a median of 128 tool calls, yet additional effort in the bash debug loop does not translate to higher score.
The bottleneck is structural rather than raw effort: of the five zero-score tasks, all produced a valid build but submitted no demo traces, leaving every rubric dimension unscored.
The write-first strategy front-loads code generation but frequently misses the demo trace delivery that closes the evaluation loop, revealing a task-completion gap orthogonal to raw coding ability.

\begin{findings}
Execution effort and tool usage volume are weak predictors of final playability; success depends on closing the full build–replay–evaluation loop.
\end{findings}

\subsection{On the Reliability of Playability Judge}
\begin{figure}[ht]
  \centering
  \includegraphics[width=0.46\linewidth]{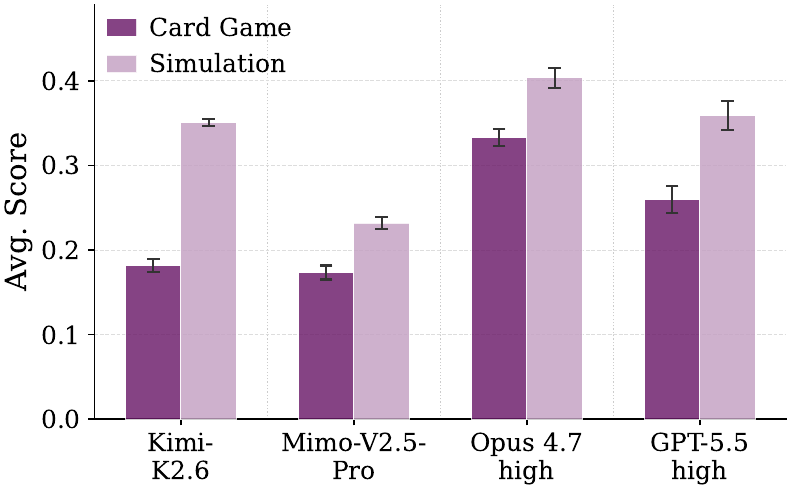}
  \caption{Judge stability on fixed gameplay evidence. Bars show mean scores over 10 repeated GPT-5.5 judge runs, and error bars show $\pm 1$ standard deviation.}
  \label{fig:judge-consistency}
\end{figure}
\paragraph{Stability: Does Fixed Evidence Receive Consistent Scores?}
\benchmark{} relies on a multimodal judge, so we first test whether identical playable evidence receives stable scores across repeated evaluations.
We fix the gameplay evidence (project, traces, videos, frames, and rubric) for tasks from two families, Simulation and Card Game, and rerun only the GPT-5.5 judge $K=10$ times.
As shown in Figure~\ref{fig:judge-consistency}, the mean scores remain consistent across repeated GPT-5.5 evaluations, and the error bars are small across the evaluated agent configurations and task families.
Quantitatively, Kimi-K2.6 exhibits a standard deviation of 0.0037 on Card Game and 0.0038 on Simulation, while Opus-4.7 high exhibits standard deviations of 0.0050 and 0.0036, respectively.
These variations are far smaller than the performance gaps across agents and families, so the observed rankings are robust to repeated judging noise.
Residual noise mainly comes from borderline visual interpretations, such as whether a card interaction or state transition is visually clear enough.

\begin{table}[ht]
  \centering
  \small
  \begin{tabular}{lrrrrrrr}
    \toprule
    Family & Human & Judge & $\Delta$M & $\Delta$D & $\Delta$V & $\Delta$A & $\Delta$Overall \\
    \midrule
    Card Game & 18.75 & 18.48 & +2.33 & +1.50 & -6.10 & -0.67 & -0.27 \\
    Idle & 32.89 & 41.65 & +6.25 & +12.19 & -3.12 & +11.50 & +8.76 \\
    Racing & 17.42 & 19.81 & -2.92 & +3.44 & -8.33 & +8.20 & +2.38 \\
    \midrule
    Overall & 22.69 & 26.02 & +1.92 & +5.38 & -5.87 & +5.80 & +3.32 \\
    \bottomrule
  \end{tabular}
  \caption{Preliminary human calibration on Kimi-K2.6 submissions from three
  families. Human and Judge report overall scores; $\Delta$ columns report
  Judge--Human differences in percentage points.}
  \label{tab:human-calibration}
\end{table}

\paragraph{Calibration: How Does the Judge Compare with Humans?}
We also run a preliminary human calibration on the same replayed gameplay evidence from three families.
As shown in Table~\ref{tab:human-calibration}, the multimodal judge is broadly aligned with human scores but slightly more permissive overall.
The main discrepancy is category-specific: human annotators are stricter on Content Depth and Art and Presentation, while the judge is stricter on Functional Visuals.
At the family level, Card Game is nearly matched, whereas Idle shows the largest permissive gap, suggesting that future calibration should focus on content variety and presentation judgments.
Because this subset is small and is not designed to estimate inter-annotator agreement, we treat it as a calibration check rather than a definitive human-agreement study.

\begin{findings}
The playability judge is stable across repeated evaluations and exhibits mild permissiveness in content and presentation dimensions.
\end{findings}

\subsection{On the Decomposability of Game Generation Ability}
\begin{figure}[ht]
  \centering
  \includegraphics[width=0.78\linewidth]{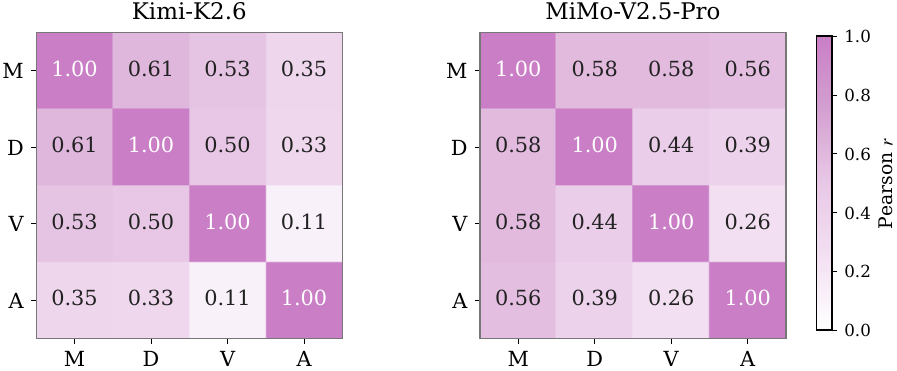}
  \caption{Correlation among rubric categories for Kimi-K2.6 and MiMo-V2.5-Pro.
  Mechanics, Depth, and Visuals are moderately coupled, while Art and
  Presentation is more weakly coupled with the other categories.}
  \label{fig:metric-correlation}
\end{figure}

We examine whether the four rubric categories rise and fall together across Kimi-K2.6 and MiMo-V2.5-Pro tasks.
As shown in Figure~\ref{fig:metric-correlation}, Core Mechanics correlates moderately with Content Depth ($r=0.61$) and Functional Visuals ($r=0.53$) for Kimi-K2.6, suggesting that stronger interaction loops often expose more game
state and feedback.
However, Art and Presentation is much less coupled with the other categories, especially Functional Visuals ($r=0.11$).
MiMo-V2.5-Pro shows a similar but more globally coupled pattern, with Art and Presentation correlating with Mechanics, Content Depth, and Functional Visuals at $r=0.56$, $r=0.39$, and $r=0.26$, respectively.
This supports the main-result observation that a mechanically recognizable game does not automatically become a complete, visually coherent, or polished game artifact.

\begin{findings}
Game generation capability is partially factorized: mechanics, content, visual feedback, and presentation are not fully coupled.
\end{findings}

\section{Related Work}
\paragraph{Coding Agents and Software Engineering Evaluation.}
Coding agents have evolved from code completion models such as CodeBERT~\citep{feng2020codebert} and StarCoder~\citep{li2023starcoder} to autonomous software engineering systems capable of repository navigation, multi-file editing, and iterative debugging. Recent frameworks, including SWE-agent~\citep{yang2024swe}, OpenHands~\citep{wang2025openhands}, ChatDev~\citep{qian2024chatdev}, and MetaGPT~\citep{hong2024metagpt}, increasingly evaluate agents on realistic software development tasks. Existing software engineering benchmarks, however, primarily assess correctness through code patches, unit tests, or issue resolution outcomes, assuming that executable behavior can be sufficiently captured by static artifacts. End-to-end game generation challenges this assumption because success depends on the quality of interactive runtime behavior rather than source code alone.

\paragraph{GUI Agents and Interactive Evaluation.}
Recent advances in GUI agents enable models to interact directly with desktop and web environments through visual perception and native interface actions. Benchmarks such as Mind2Web~\citep{deng2023mind2web}, WebArena~\citep{zhou2024webarena}, and OSWorld~\citep{xie2024osworld} emphasize end-to-end task completion under interactive settings, while systems including CogAgent~\citep{hong2024cogagent} and OSAtlas~\citep{wu2025atlas} demonstrate increasingly capable computer-use behaviors. These developments highlight the importance of evaluating agents through interaction rather than static outputs alone. However, existing GUI benchmarks focus on accomplishing predefined interface tasks rather than synthesizing executable software artifacts such as games.

\paragraph{Game Generation Benchmarks.}
Recent work has begun to evaluate coding agents on game generation, but existing benchmarks cover different parts of the evaluation contract.
OpenGame-Bench~\citep{jiang2026opengame} asks agents to generate complete games from open-ended prompts, but it targets web games and relies primarily on static or page-level judgment rather than gameplay interaction.
GameDevBench~\citep{chi2026gamedevbench} brings evaluation into Godot, but studies localized tutorial-derived edits within existing projects rather than end-to-end construction of complete games.
Closest to our setting, WebGameBench~\citep{zhang2026webgamebench} evaluates delivered browser-native games through real browser interaction; however, it remains outside engine-native game development and depends on evaluation-side exploration to discover gameplay behavior.
Concurrent work on GameGen-Verifier~\citep{jia2026gamegen} improves runtime verification through state injection, but does not evaluate open-ended game construction from specifications.
In contrast, \benchmark{} jointly requires agents to build complete Godot projects and to provide replayable demonstrations, enabling interaction-grounded evaluation without turning verification itself into an autonomous game-exploration problem.

\section{Conclusion}
We presented an interaction-grounded evaluation framework for end-to-end game generation and instantiate it as \benchmark{}, a Godot-based benchmark for coding agents.
The central claim is that game generation cannot be evaluated as ordinary code synthesis: a benchmark must preserve the engine environment, require a complete artifact, and judge the artifact through interaction.
Our results show that current frontier agents remain far from reliable under this contract.
They can often produce recognizable mechanics or runnable prototypes, but still struggle to assemble coherent games with sufficient content, functional feedback, presentation quality, and completed demonstration traces.
More broadly, our findings suggest that evaluating executable creative software systems requires interaction-grounded assessment beyond static code inspection, build success, or visual plausibility alone.

\section*{Acknowledgment}
This work was supported by Major Frontier Exploration Program (Grant No. C10120250085) from the Shenzhen Medical Academy of Research and Translation (SMART),  Shenzhen Medical Research Fund (B2503005),  NSFC grant 72495131, the 1+1+1 CUHK-CUHK(SZ)-GDSTC Joint Collaboration Fund, Guangdong Provincial Key Laboratory of Mathematical Foundations for Artificial Intelligence (2023B1212010001), and the International Science and Technology Cooperation Center, Ministry of Science and Technology of China (under grant 2024YFE0203000).

\section*{Limitations}
\benchmark{} has several scope limitations.
It focuses on 2D game generation in Godot, which makes the benchmark lightweight and reproducible for headless evaluation, but does not cover other major engines such as Unity and Unreal.
It also does not evaluate 3D games, multiplayer systems, large-scale physics, or long-form production workflows.
In addition, the current verifier scores visual gameplay evidence only; audio-dependent aspects of rhythm, horror, or sports games are represented through visual behavior rather than direct audio evaluation.

\benchmark{} also has evaluation limitations.
It relies on a multimodal judge to score replayed gameplay evidence, so results may still be affected by judge-model bias, API drift, or limitations in visual understanding.
Finally, \benchmark{} does not measure whether a generated game is subjectively fun.
Instead, it measures whether the agent follows the game specification and realizes the requested mechanics, content, visual state, and presentation in an executable artifact.

\bibliographystyle{unsrt}
\bibliography{refs}

\newpage
\appendix

\section{Evaluation Details}\label{app:evaluation-details}

\paragraph{Runtime Environment.} \benchmark is implemented on Harbor~\cite{Harbor_Framework} and runs each trial in a local subprocess sandbox.
The reference setup uses Ubuntu 22.04, Python 3.12, and Godot 4.6.2 stable.
The verifier depends on Xvfb for headless display, xdotool for synthetic input, and ffmpeg for screen recording and frame extraction.
Each task uses an agent timeout of 7200 seconds, a verifier timeout of 1800 seconds.
We do not impose an explicit tool-call limit beyond the task timeout.
The agent is given a fixed workspace rooted at \texttt{/workspace/} and must place the generated Godot project at \texttt{/workspace/game}.

\paragraph{Submission Format.}
Each submission consists of a Godot project and one to ten demonstration trace files named \texttt{*.json} under \texttt{/workspace/game/demo\_outputs/}.
Only the first ten traces by filename are evaluated.
A trace contains an optional scenario identifier, a total replay duration in frames, and a time-ordered list of mouse or keyboard events in a fixed 1280$\times$720 viewport.
Demos can deterministically initialize a requested
state such as a battle, upgrade screen, or late-game setup.
The trace schema is:
\begin{prompt}{Demo Trace Schema}
\begin{verbatim}
{
  "scenario": "title_flow",
  "duration_frames": 360,
  "events": [
    {"frame": 30,  "type": "mouse_click",
     "button": "left", "x": 300, "y": 360},
    {"frame": 90,  "type": "key_press", "keycode": "1"},
    {"frame": 180, "type": "key_press", "keycode": "SPACE"},
    {"frame": 300, "type": "wait"}
  ]
}
\end{verbatim}
\end{prompt}

\paragraph{Replay and Judge.}
For each valid trace, the verifier starts a fresh Godot instance, replays input at 30 frames per second, records the 1280$\times$720 display through ffmpeg, and stores a compressed 854$\times$480 version.
The verifier samples frames every 0.5 seconds, corresponding to two frames per second.
Each rubric may cap the number of demos and the sampled time window; the default benchmark configuration uses at most ten demos and samples at most a deterministic 20-second window per demo.
We use \texttt{GPT-5.5} as the multimodal judge.
The judge prompt template is shown below:

\begin{prompt}{Judge Prompt Template}
You are a strict but fair video-game evaluator. Given a short playthrough recording of a Godot 2D game, decide how clearly each listed requirement is demonstrated. Score every requirement on a 0.0 to 1.0 scale where 0.0 = not demonstrated at all (or contradicted), 0.5 = partially demonstrated / ambiguous, and 1.0 = clearly and unambiguously demonstrated by what is visible in the recording. Reply with strict JSON only, no prose, no markdown, no code fences.\\

Evaluate the recording against each of the following requirements.\\

Requirements:

\{requirements\}\\

Return JSON in exactly this shape (no extra keys, no markdown):
\begin{verbatim}
{
  "scores": {"<requirement_id>": <0..1>},
  "rationales": {"<requirement_id>": "<one short sentence>"}
}
\end{verbatim}
\end{prompt}

\section{Full Family Results}\label{app:family-results}

{
\scriptsize
\setlength{\tabcolsep}{2.3pt}
\renewcommand{\arraystretch}{0.9}

\newcommand{\ocell}[1]{\cellcolor[HTML]{f4f0fa}#1}
\newcommand{\agentrow}[1]{%
  \rowcolor{black!8}\multicolumn{18}{c}{\textbf{#1}} \\
}
\begin{table}[ht]
  \centering
  \resizebox{\linewidth}{!}{%
  \begin{tabular}{ll*{16}{c}}
    \toprule
    \textbf{Model} & \textbf{Metric ($\uparrow$)}
    & \textbf{Overall}
    & \textbf{Platformer}
    & \textbf{Strategy}
    & \textbf{Tycoon}
    & \textbf{Open-world}
    & \textbf{Roguelike}
    & \textbf{Visual novel}
    & \textbf{Puzzle}
    & \textbf{Shooter}
    & \textbf{Simulation}
    & \textbf{Card game}
    & \textbf{Horror}
    & \textbf{Rhythm}
    & \textbf{Idle}
    & \textbf{Racing}
    & \textbf{Sports} \\
    \midrule
    \agentrow{Claude Code}
    \multirow{5}{*}{\modellogo{figures/anthropic.png}\,Opus-4.7 high}
      & M & \textbf{55.34} & \underline{45.77} & \textbf{54.91} & \textbf{48.31} & \underline{47.88} & \textbf{64.87} & \textbf{56.33} & \underline{51.43} & \underline{55.00} & \textbf{61.17} & \textbf{54.67} & \textbf{72.67} & \textbf{62.00} & \underline{76.67} & \underline{55.83} & \textbf{66.25} \\
      & D & \textbf{39.48} & \textbf{35.50} & \textbf{46.78} & \textbf{39.42} & \underline{34.43} & \textbf{38.79} & \underline{42.22} & \underline{26.43} & \textbf{40.57} & 37.33 & \textbf{26.25} & \textbf{49.00} & \underline{40.00} & \underline{61.25} & \textbf{45.31} & 40.62 \\
      & V & \textbf{42.78} & \textbf{46.28} & \textbf{62.60} & \textbf{45.11} & \textbf{29.57} & \textbf{46.50} & \textbf{50.99} & \underline{43.36} & \underline{38.20} & \textbf{44.13} & \textbf{18.48} & \textbf{50.50} & \underline{24.70} & \underline{23.12} & \underline{23.15} & \textbf{39.38} \\
      & A & \textbf{36.86} & \underline{29.55} & \textbf{30.65} & \textbf{24.69} & \textbf{48.94} & \textbf{28.05} & \textbf{35.21} & \underline{35.44} & 34.58 & \underline{26.93} & \textbf{38.92} & \textbf{61.92} & \textbf{55.90} & \underline{58.82} & \textbf{50.15} & \textbf{51.95} \\
      & \ocell{\textbf{Overall}} & \ocell{\textbf{41.46}} & \ocell{\underline{36.57}} & \ocell{\textbf{44.73}} & \ocell{\textbf{36.45}} & \ocell{\textbf{40.80}} & \ocell{\textbf{40.10}} & \ocell{\textbf{43.60}} & \ocell{\underline{35.87}} & \ocell{\underline{40.28}} & \ocell{\underline{38.29}} & \ocell{\textbf{33.78}} & \ocell{\textbf{57.30}} & \ocell{\textbf{46.57}} & \ocell{\underline{56.99}} & \ocell{\textbf{45.26}} & \ocell{\underline{48.24}} \\
    \multirow{5}{*}{\modellogo{figures/xiaomimimo.png}\,MiMo-V2.5-Pro}
      & M & 32.33 & 32.58 & 28.81 & 41.34 & 17.22 & 34.13 & 27.64 & 33.12 & 40.29 & 37.17 & 26.67 & 45.67 & 31.00 & 50.83 & 23.75 & 32.92 \\
      & D & 22.59 & 17.50 & 20.11 & 36.31 & 17.08 & 21.27 & 21.02 & 21.09 & 19.00 & 28.17 & 12.25 & 26.00 & 13.75 & 41.25 & 16.56 & 38.75 \\
      & V & 27.45 & 34.41 & 29.41 & 39.42 & 11.81 & 35.85 & 29.63 & 32.02 & 30.10 & 30.59 & 4.95 & 28.00 & 16.34 & 11.88 & 9.51 & 19.38 \\
      & A & 20.65 & 18.46 & 14.12 & 19.53 & 22.96 & 18.19 & 15.37 & 21.21 & 25.02 & 20.53 & 14.59 & 41.14 & 16.69 & 42.94 & 24.57 & 29.29 \\
      & \ocell{\textbf{Overall}} & \ocell{24.10} & \ocell{22.64} & \ocell{20.72} & \ocell{\underline{31.66}} & \ocell{18.37} & \ocell{24.31} & \ocell{21.36} & \ocell{24.58} & \ocell{25.96} & \ocell{27.21} & \ocell{14.14} & \ocell{34.55} & \ocell{17.76} & \ocell{38.87} & \ocell{19.39} & \ocell{31.66} \\
    \midrule
    \agentrow{Codex}
    \multirow{5}{*}{\modellogo{figures/openai.png}\,GPT-5.5 high}
      & M & \underline{54.36} & \textbf{50.51} & \underline{50.96} & \underline{42.64} & \textbf{54.48} & 50.99 & \underline{56.00} & \textbf{67.97} & \textbf{64.00} & \underline{53.83} & \underline{50.67} & \underline{59.67} & \underline{57.33} & \textbf{90.83} & \textbf{57.50} & \underline{52.08} \\
      & D & \underline{38.61} & \underline{34.66} & \underline{40.69} & \underline{36.85} & \textbf{39.23} & 32.60 & \textbf{45.61} & \textbf{37.97} & \underline{33.71} & \underline{37.50} & \underline{23.25} & \underline{36.50} & \textbf{40.75} & \textbf{72.19} & \underline{35.00} & \textbf{55.62} \\
      & V & \underline{41.84} & \underline{45.49} & \underline{56.61} & \underline{41.01} & \underline{27.29} & \underline{43.42} & \underline{50.46} & \textbf{53.34} & \textbf{44.89} & \underline{40.10} & \underline{17.73} & \underline{44.00} & \textbf{32.84} & \textbf{28.12} & \textbf{25.23} & \underline{33.75} \\
      & A & \underline{32.94} & \textbf{30.65} & \underline{26.94} & 17.45 & \underline{40.51} & 25.65 & \underline{28.50} & \textbf{43.75} & \textbf{37.62} & \textbf{33.11} & 20.29 & \underline{59.20} & \underline{42.45} & \textbf{62.29} & 35.69 & \underline{49.61} \\
      & \ocell{\textbf{Overall}} & \ocell{\underline{39.49}} & \ocell{\textbf{37.26}} & \ocell{\underline{39.81}} & \ocell{31.55} & \ocell{\underline{40.17}} & \ocell{34.55} & \ocell{\underline{42.15}} & \ocell{\textbf{46.80}} & \ocell{\textbf{41.30}} & \ocell{\textbf{38.81}} & \ocell{\underline{25.50}} & \ocell{\underline{49.05}} & \ocell{\underline{42.65}} & \ocell{\textbf{64.91}} & \ocell{\underline{37.15}} & \ocell{\textbf{49.71}} \\
    \multirow{5}{*}{\modellogo{figures/deepseek.png}\,DeepSeek-V4-Pro}
      & M & 2.25 & 0.39 & 1.35 & 0.00 & 0.92 & 1.71 & 0.45 & 0.00 & 0.00 & 6.50 & 0.00 & 7.33 & 0.00 & 33.33 & 0.00 & 8.33 \\
      & D & 1.69 & 1.00 & 0.35 & 0.00 & 0.00 & 1.00 & 0.82 & 0.00 & 0.00 & 6.33 & 0.00 & 5.00 & 0.00 & 13.44 & 0.00 & 18.12 \\
      & V & 1.97 & 2.67 & 0.88 & 0.00 & 1.11 & 3.48 & 2.56 & 0.00 & 0.00 & 5.62 & 0.00 & 2.00 & 0.00 & 6.88 & 0.00 & 11.25 \\
      & A & 2.63 & 2.73 & 0.59 & 0.00 & 0.28 & 2.75 & 0.91 & 0.00 & 0.00 & 4.93 & 7.00 & 10.40 & 4.83 & 17.92 & 0.00 & 10.38 \\
      & \ocell{\textbf{Overall}} & \ocell{2.15} & \ocell{1.77} & \ocell{0.66} & \ocell{0.00} & \ocell{0.40} & \ocell{2.09} & \ocell{1.06} & \ocell{0.00} & \ocell{0.00} & \ocell{5.76} & \ocell{2.45} & \ocell{6.79} & \ocell{1.69} & \ocell{17.01} & \ocell{0.00} & \ocell{12.91} \\
    \midrule
    \agentrow{Kimi Code}
    \multirow{5}{*}{\modellogo{figures/kimi.jpeg}\,Kimi-K2.6}
      & M & 39.76 & 34.32 & 47.65 & 38.50 & 26.99 & \underline{52.31} & 45.73 & 30.47 & 43.00 & 47.50 & 29.00 & 34.67 & 41.00 & 64.58 & 20.00 & 39.17 \\
      & D & 28.07 & 22.71 & 32.22 & 32.04 & 23.10 & \underline{33.22} & 40.09 & 18.12 & 30.43 & \textbf{40.83} & 11.50 & 17.50 & 14.50 & 38.75 & 8.12 & \underline{44.38} \\
      & V & 33.66 & 39.27 & 44.17 & 39.30 & 19.05 & 43.30 & 46.49 & 35.51 & 29.65 & 38.42 & 10.57 & 19.00 & 22.91 & 15.62 & 8.33 & 25.62 \\
      & A & 27.99 & 28.43 & 21.23 & \underline{22.60} & 24.62 & \underline{26.27} & 27.75 & 26.23 & \underline{34.61} & 25.60 & \underline{24.33} & 42.19 & 31.84 & 45.87 & \underline{36.32} & 46.74 \\
      & \ocell{\textbf{Overall}} & \ocell{30.65} & \ocell{28.94} & \ocell{32.48} & \ocell{30.79} & \ocell{23.61} & \ocell{\underline{35.17}} & \ocell{37.85} & \ocell{25.42} & \ocell{33.66} & \ocell{36.14} & \ocell{18.48} & \ocell{28.94} & \ocell{25.80} & \ocell{41.65} & \ocell{19.81} & \ocell{41.61} \\
    \midrule
    \agentrow{Code Buddy}
    \multirow{5}{*}{\modellogo{figures/glm.png}\,GLM-5.1}
      & M & 25.23 & 22.37 & 19.12 & 27.25 & 22.53 & 30.09 & 21.45 & 35.62 & 24.29 & 42.67 & 18.67 & 25.00 & 15.33 & 40.00 & 33.33 & 12.92 \\
      & D & 17.80 & 11.95 & 18.97 & 22.21 & 16.90 & 19.06 & 16.27 & 20.94 & 16.29 & 30.00 & 14.25 & 10.50 & 14.25 & 24.06 & 8.12 & 25.62 \\
      & V & 21.14 & 22.35 & 25.85 & 26.40 & 11.75 & 25.12 & 24.33 & 29.49 & 20.37 & 34.58 & 7.91 & 13.00 & 12.13 & 11.25 & 7.86 & 12.50 \\
      & A & 14.59 & 11.99 & 9.19 & 9.01 & 17.26 & 13.35 & 12.26 & 18.91 & 15.26 & 20.17 & 14.01 & 36.68 & 11.76 & 26.57 & 14.05 & 19.97 \\
      & \ocell{\textbf{Overall}} & \ocell{18.29} & \ocell{15.09} & \ocell{16.60} & \ocell{18.98} & \ocell{17.10} & \ocell{19.62} & \ocell{16.88} & \ocell{23.71} & \ocell{17.74} & \ocell{29.15} & \ocell{13.88} & \ocell{22.21} & \ocell{13.22} & \ocell{25.41} & \ocell{13.94} & \ocell{19.77} \\
    \multirow{5}{*}{\modellogo{figures/minimax.jpeg}\,MiniMax-M2.7}
      & M & 14.27 & 9.96 & 13.24 & 20.69 & 2.98 & 21.00 & 12.91 & 4.53 & 21.14 & 30.50 & 16.33 & 17.67 & 9.00 & 22.92 & 14.58 & 10.00 \\
      & D & 9.92 & 5.00 & 10.51 & 14.56 & 4.98 & 11.72 & 12.18 & 5.78 & 11.86 & 20.17 & 3.00 & 12.25 & 11.00 & 11.88 & 9.69 & 10.31 \\
      & V & 14.92 & 15.87 & 17.84 & 25.63 & 6.67 & 20.48 & 16.89 & 16.51 & 17.15 & 29.12 & 1.20 & 2.00 & 4.67 & 3.75 & 2.60 & 2.50 \\
      & A & 8.85 & 6.37 & 4.13 & 7.48 & 9.34 & 6.84 & 5.77 & 3.04 & 13.29 & 15.25 & 10.31 & 25.19 & 14.24 & 21.79 & 12.00 & 8.94 \\
      & \ocell{\textbf{Overall}} & \ocell{10.95} & \ocell{7.85} & \ocell{9.79} & \ocell{14.66} & \ocell{6.46} & \ocell{12.72} & \ocell{10.82} & \ocell{6.24} & \ocell{14.55} & \ocell{21.34} & \ocell{7.29} & \ocell{16.05} & \ocell{10.88} & \ocell{15.78} & \ocell{10.17} & \ocell{8.61} \\
    \bottomrule
  \end{tabular}%
  }
  \caption{Family-level and benchmark-level results (\%). For each model, rows report
  core mechanics (M), content depth (D), functional visuals (V), art and
  presentation (A), and overall score.}
  \label{tab:family-results}
\end{table}
}

\newpage
\section{Case Study}
\label{app:case-study}

\begin{figure}[ht]
  \centering
  \includegraphics[width=0.76\linewidth]{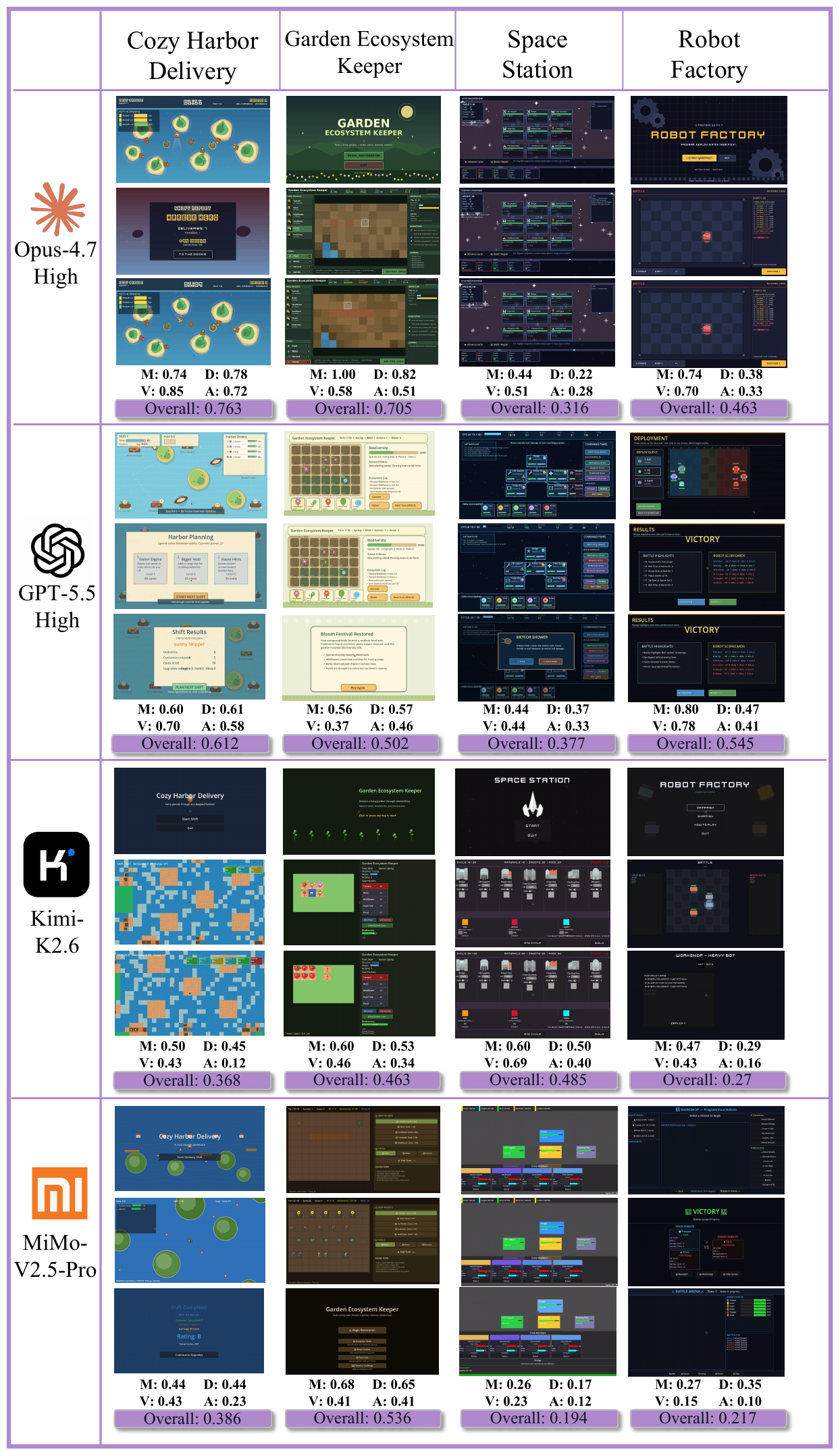}
  \caption{Case Study: Four Models on Four Representative Tasks. Each cell shows three sampled gameplay frames from the agent's submitted demo traces, together with per-category scores (\textit{M}: Core Mechanics, \textit{D}: Content Depth, \textit{V}: Functional Visuals, \textit{A}: Art \& Presentation), and the overall score shown below.}
  \label{fig:case-study-by-diff-models}
\end{figure}

\begin{prompt}{Example \texttt{rubric.json} of Strategy-Skirmish}
\begin{verbatim}
  "score_formula": "BUILD * (0.15*((M1+M2+M3+M4)/4) + 0.35*((D1+D2+D3+D4)/4)
             + 0.15*((V1+V2+V3+V4)/4) + 0.35*((A1+A2+A3+A4)/4))",
  "max_demos": 10, "max_demo_seconds": 20,
  "build_check": {
  "id": "BUILD", "cmd": "godot --headless --path /workspace/game --quit-after 5" },
  "categories": [
    {"name": "Core Mechanics", "items": ["M1", "M2", "M3", "M4"]},
    {"name": "Content Depth", "items": ["D1", "D2", "D3", "D4"]},
    {"name": "Functional Visuals", "items": ["V1", "V2", "V3", "V4"]},
    {"name": "Art & Presentation", "items": ["A1", "A2", "A3", "A4"]} ],
  "requirements": [
    {
      "id": "M1", "agg": "max",
      "description": "The irreducible tactical loop is playable: the player can select a 
      friendly unit, see a limited legal movement range highlighted on the grid, choose a 
      legal destination, and watch the unit move to the chosen cell. Illegal clicks 
      outside the range are rejected without corrupting state. Selecting a different unit 
      or clicking the same unit again cancels the current selection. Score 0 if selecting 
      and moving a unit is missing, scripted-only, or not controlled by player input. 
      Score 1 requires the mechanic to be presented in a visually authored context with 
      real sprites or illustrated assets. Score 0.5 at most if the mechanic works but is 
      represented entirely by programmatic shapes, solid-color fills, or default Godot widgets."
    },
    ...
    {
      "id": "D1", "agg": "max",
      "description": "The player's squad has meaningful depth: at least three controllable unit
      types or named characters are present, visually identifiable, and functionally different
      through HP, movement range, attack range, damage, or special ability. Score 1 only if the
      differences change tactical choices during battle. Score 0 if there is only one player
      unit type or if all units share identical stats and behavior. Score 0.5 at most if
      only labels or decorative differences distinguish units with no observable gameplay impact.
      Score 1 requires content variants to be visually distinct with real authored art (different
      sprites, illustrated elements). Score 0.5 at most if variants exist but are represented by
      programmatic shapes, color swaps, or text labels without real asset differences."
    },
    ...
    {
      "id": "V1", "agg": "mean",
      "description": "Battle readability is strong throughout gameplay: grid cells, walkable
      terrain, blocked terrain, friendly units, enemy units, selected unit, legal move 
      cells, and attackable targets are all visually distinct at 1280x720. Score 1 only 
      if a player can understand legal actions without guessing. Score 0 if key states 
      blend together, overlays are missing, or placeholder shapes make sides and zones 
      ambiguous. Score 1 requires the information to be presented through themed, styled 
      UI with real assets. Score 0.5 at most if readability relies on plain text, 
      unstyled labels, or raw geometric shapes."
    },
    ...
    {
      "id": "A1", "agg": "mean",
      "description": "The game has a coherent dark-fantasy art direction: terrain, units, UI,
      effects, and backgrounds share one consistent visual language with a deliberate 
      limited palette and high contrast. Score 1 only if the presentation looks like an 
      authored vertical slice rather than a prototype. Score 0 if visible gameplay relies 
      on raw ColorRect shapes, default Godot widgets, mismatched asset packs, or bright 
      off-brief placeholder colors."
    },
    ...
  ]
\end{verbatim}
\end{prompt}

\begin{prompt}{Example \texttt{instruction.md} of Strategy-Skirmish}
\begin{verbatim}
# Strategy: Skirmish

Build a **dark-fantasy tactical skirmish** in Godot 4 at `/workspace/game/`.
This is not a prototype. It is a **complete, shippable micro-game** that could
sit on an itch.io page or Steam as a polished vertical slice.

## Core Vision

The player commands a small, outnumbered squad through desperate grid-based
battles where every move is a commitment and every loss is permanent for the
fight. The fantasy is **grim tactical survival** — a handful of specialists
against a tide of enemies, where positioning is life and a single misread costs
a unit you cannot replace mid-battle. The tone channels *Into the Breach* meets
*Darkest Dungeon* at a smaller scale: limited palette, high contrast, tense
decisions. The best version makes the player feel like a cornered general
finding the one sequence of moves that turns impossible odds into a narrow
victory.

## What the Player Experiences

A moody title screen sets the dark-fantasy tone immediately. The player begins
and receives a brief tactical briefing — the squad's objective, the threat
ahead, the stakes — before the grid appears.

The battle is turn-based and deliberate. The player selects a unit, sees its
limited movement range light up on the grid, and commits it to a position.
Enemies are visible, aggressive, and numerous — the squad is always
outnumbered. After the player spends their actions, an End Turn command hands
control to the enemy, which advances with purpose: flanking, closing distance,
attacking when in range. Then control returns and the cycle repeats.

Combat is lethal and readable. Attacks require proximity or a clear range
indicator, reduce persistent HP, and kill. Dead units vanish from the board and
stop blocking or threatening. The player's squad members are specialists —
different movement ranges, attack patterns, HP pools, or abilities — so
choosing who moves where and who attacks what is the core decision space.

The battlefield itself adds tactical texture: terrain obstacles funnel movement,
hazards punish careless positioning, or objectives create time pressure beyond
simple elimination. Multiple battle layouts keep the experience from feeling
solved after one fight.

Victory comes from eliminating all enemies; defeat from losing the squad. Either
outcome lands on a styled result screen showing what happened, and the player
can retry or return to the title without relaunching. The entire arc — title,
briefing, battle, result — flows as one continuous authored experience.

{Resource Instruction}

{Foramt Instruction}
\end{verbatim}
\end{prompt}

\begin{prompt}{Resource Instruction in \texttt{instruction.md}}
\begin{verbatim}
## Assets

2D assets are mounted read-only at:

- `/workspace/assets/library/` — Kenney CC0 packs (sprites, tiles, UI, fonts).
- `/workspace/assets/library-oga/` — OpenGameArt entries; respect each
  subdir's `LICENSE.txt`.

Browse the library and choose packs.
Copy what you need into your project's `assets/` folder.

## Project layout

```
/workspace/game/
  project.godot
  Main.tscn
  demo_outputs/    ← your input traces (1–10 files)
  scripts/  scenes/  assets/
```

The build must launch cleanly with:

```
godot --headless --path /workspace/game --quit-after 5
```

A reference for Godot CLI flags is at `/workspace/tools/godot_command_line.md`.
**Engine flags like `--headless` and `--quit-after N` must come BEFORE `--`** —
anything after `--` is forwarded to the project as user args and silently
ignored by the engine. Correct shape:
`godot --headless --quit-after 5 --path . -- --scenario near_victory`.

A screenshot helper is available at `/workspace/tools/screenshot.sh`.
Use it to actually see what your UI / battlefield /
result screens look like.

```
/workspace/tools/screenshot.sh --path /workspace/game \
      -- --out /workspace/frame.png --frames 60
```

To screenshot a specific scenario, append `--scenario <id>` after `--`. The
helper consumes only `--out` / `--frames` / `--scene`; remaining args stay in
`OS.get_cmdline_user_args()` for your game code to read. Example:

```
/workspace/tools/screenshot.sh --path /workspace/game \
      -- --out /workspace/battle_debug.png --frames 120 --scenario battle
```
\end{verbatim}
\end{prompt}

\begin{prompt}{Format Instruction in \texttt{instruction.md}}
\begin{verbatim}
## Demos

Ship **1–10 input-trace files** under `/workspace/game/demo_outputs/`, one per
demo, each named `*.json`. The evaluator launches a fresh game per trace,
replays your trace as synthetic mouse and keyboard input at 1280×720, and
records the screen. Only the first 10 traces by filename are evaluated;
recordings longer than 20 s are sampled from a random 20 s window.

### Scenarios

Normal play should start from the title screen and demonstrate the task's
core gameplay loop.
Demo playback must be deterministic. For demos that need a specific state
(a specific level, combat state, upgrade screen, result state, or late-game
setup), define named scenarios your game loads when launched with:

```
godot --path /workspace/game -- --scenario <id>
```

When `--scenario <id>` is present the game must skip menus, set up the named
state deterministically (seed any RNG), and begin accepting input immediately.

### Trace file format

```json
{
  "scenario": "title_flow",
  "duration_frames": 360,
  "events": [
    {"frame": 30,  "type": "mouse_click", "button": "left", "x": 300, "y": 360},
    {"frame": 90,  "type": "key_press",   "keycode": "1"},
    {"frame": 180, "type": "key_press",   "keycode": "SPACE"},
    {"frame": 300, "type": "wait"}
  ]
}
```

- `scenario` — optional; omit for a normal game launch from the title screen.
- `duration_frames` — total frames to record at 30 fps; cap at **600 (20 s)**.
- `events` — time-ordered inputs. Coordinates are pixels in the 1280×720
  viewport. Supported types:
  - `mouse_click`: `{frame, type, button: "left"|"right", x, y}`
  - `mouse_down` / `mouse_up`: `{frame, type, button: "left"|"right", x, y}` —
    use these for drag interactions: emit `mouse_down` at the start point,
    one or more `mouse_move` events along the way, and `mouse_up` at the end.
    A `mouse_click` is a `mouse_down` + `mouse_up` at the same point in tight
    succession.
  - `mouse_move`: `{frame, type, x, y}`
  - `key_press` / `key_down` / `key_up`: `{frame, type, keycode}` — keycodes:
    `A`–`Z`, `0`–`9`, `ESCAPE`, `ENTER`, `SPACE`, `TAB`, `BACKSPACE`,
    `DELETE`, `SHIFT`, `CTRL`, `ALT`, `UP`, `DOWN`, `LEFT`, `RIGHT`.
  - `wait`: `{frame, type}` — anchor frame, no input.

Replay must be deterministic: same trace, fresh launch, same outcome every time.
\end{verbatim}
\end{prompt}

\end{document}